# Parameterization method of reservoir properties for ensemble-based data assimilation using intermediate latent space of StyleGAN

Marcio A. Sampaio[a1], Paulo H. Ranazzi[a2], Martin J. Blunt[b3]

[a]Departamento de Engenharia de Minas e de Petróleo, Escola Politécnica, Universidade de São Paulo, 05508-030, SP, Brasil.

[b]Department of Earth Science and Engineering, Imperial College London, South Kensington, London, UK

[1]Corresponding author. E-mail address: marciosampaio@usp.br (Marcio A. Sampaio). ORCID: 0000-0003-1125-7218

[2]Contributing author. E-mail address: ranazzi@usp.br (Paulo H. Ranazzi). ORCID: 0000-0002-4515-4797

[3]Contributing author. E-mail address: m.blunt@imperial.ac.uk (Martin J. Blunt). ORCID: 0000-0002-8725-0250

## Authorship contribution statement

Author 1: Conceptualization, Methodology, Data Curation, Formal analysis, Investigation, Software, Validation, Visualization, Writing. Author 2: Data curation, Datasets creation, Investigation, Validation, Writing. Author 3: Conceptualization, Data curation, Investigation, Validation, Writing, Visualization, Supervision.

## ABSTRACT

Ensemble smoothers are the most successful and efficient techniques currently available for history matching. However, because these methods rely on Gaussian assumptions, their performance is severely degraded when the prior geology is described in terms of complex facies distributions (non-Gaussian). In this way, for these methods, we need to apply efficient parameterization techniques. Currently, the most efficient methods for performing parameterization are deep learning models. However, given the variety of existing deep learning models, studies have not identified which is most suitable for use with ensemble-based methods, although some important models had already been evaluated. Therefore, trying to fill this gap, this study selected the three leading models found in the literature to determine the best method to employ, highlighting the advantages and disadvantages of each one. Based on a recent literature review, the most promising models selected were VAE-GAN, Latent Diffusion, and StyleGAN models. As a novel aspect of this work, data assimilation with the second generation of StyleGAN (StyleGAN2) model was performed using the latent z-space and intermediate w-space, separately. They were applied in two 2D case studies: one categorical (three facies) and the other continuous. The results demonstrated that all three models are highly efficient, with the StyleGAN2 model standing out for generating samples with geological realism and achieving excellent data matching in the cases studied. Our findings show that performing data assimilation with StyleGAN2 using the intermediate space (w-space) yielded better results than the traditional application in the latent space (z-space). This is due to the fact that ESMDA uses linear updates and the w-space is much more linear and disentangled than the highly entangled z-space, thereby ensuring that the updated vectors remain close to realistic geological patterns. These results were validated using main geostatistical and history matching metrics.

# 1. INTRODUCTION

Ensemble-based methods represent the state-of-the-art for data assimilation (DA) application for history matching of the reservoir models. The main disadvantages of ensemble-based methods are their poor performance in highly nonlinear models, spurious correlations from limited ensemble size, and reliance on Gaussian assumptions, which degrades performance when reservoir prior parameters exhibit non-Gaussian distributions. Among ensemble-based methods, the Ensemble Smoother with Multiple Data Assimilation (ESMDA), proposed by Emerick and Reynolds (2013), is one of the most widely used. For a comprehensive review of the ensemble-based methods, readers are referred to Aanonsen et al. (2009) and Oliver and Chen (2011). Despite these methods have been applied successfully, they sometimes fail to preserve the geological realism of the model in reservoirs with complex facies distributions. This occurs mainly because of the underlying Gaussian assumptions on model parameters that are inherent in these methods. This fact has encouraged an intense research activity to develop Gaussian parameterizations in a latent space that maps into geologically-realistic facies realizations. Despite the large number of methods employed in the literature, conventional methods like Level Set (Chang et al., 2010; Luo et al., 2007, 2008), Truncated Pluri-Gaussian (Beucher and Renard, 2016; Silva and Deutsch, 2017), Distance Transform (Hakim-Elahi and Jafarpour, 2017), and Normal-Score Transform (Li et al., 2018) tend toward multi-Gaussian assumptions and fail to adapt to the elaborate spatial statistics required in the geological models (Ling and Jafarpour, 2024). Recent applications of deep learning have attracted attention due to the quality of the results generated. Next, we highlight the key studies that employed deep learning models in the parameterization problem to be applied in the DA of reservoir models.

Laloy et al. (2017) used a Variational Autoencoder (VAE) to construct a low-dimensional parameterization of binary facies models for data assimilation with Markov Chain

Monte Carlo (MCMC). The results showed that the dimensionality reduction (DR) approach outperforms traditional techniques like principal component analysis (PCA), optimization-PCA (OPCA) and discrete cosine transform (DCT). Two synthetic cases were used to illustrate the effectiveness of the proposed DR-based probabilistic inversion in relation to traditional approaches mentioned before. Later (Laloy et al., 2018), modified the spatial GAN to deal with high dimensional inversion problems. Synthetic 3D cases were tested to prove that GAN could capture the statistical features using low-dimensional latent variables and could perform well for inversion of channel structures.

Canchumuni et al. (2017) used an autoencoder to parameterize binary facies values in terms of continuous variables for history matching with an ensemble smoother. In other work, Canchumuni et al. (2019a) extended the same parameterization using deep belief networks (DBN), which is able to map discrete facies into continuous parameters and reconstruct back to discrete facies. The results showed that the training step of the DBN was very successful in terms of reconstructing the input facies realizations of the validation sets, but the same performance was not observed when we updated the latent vectors of the DBN with ESMDA. As previous work of these authors was based on fully-connected layers, making the computational requirements for training infeasible in practice, Canchumuni et al. (2019b) applied a convolutional variational autoencoder (CVAE) and the ESMDA to investigate the parameterization in three synthetic history-matching problems with channelized facies, two 2D cases and one 3D. The results proved promising compared to previous methods, generating well-defined channelized facies. The proposed procedure outperformed previous results obtained with standard ESMDA, ESMDA with OPCA and DBN parameterizations in terms of the reconstruction of the channel facies. However, this work highlighted the need to improve the reconstruction accuracy, especially in three-dimensional cases, and reduce the computational cost in the training process of models. Results with VAEs alone showed that are generally superior to standard PCA for non-Gaussian models, but they can produce blurry output of lower quality than GANs, or they may display unrealistic geometries. After, (Canchumuni et al., 2021) applied nine different formulations, including VAE, generative adversarial network (GAN), Wasserstein GAN (WGAN), WGAN with gradient penalty, WGAN with spectral normalization, variational auto-encoding GAN, principal component analysis (PCA) with cycle GAN, PCA with transfer style network, and VAE with style loss in the same channelized facies cases of the previous studies. They also proposed two strategies to allow the use of distance-based localization with the deep learning parameterizations. The results showed that all networks were able to generate realistic facies realizations with well-defined channels, but the DA for water cut values were not so good like images generation. Both localization strategies used in this work were able to solve the ensemble collapse problem, especially found in the VAE model application.

Liu and Durlofsky (2021) propose the 3D CNN-PCA procedure for geological parameterization. This work introduced a new supervised learning-based reconstruction loss, which is used in combination with style loss and hard data loss to handle complex 3D geomodels. The deep learning model was used as a post-processor for PCA, which is a traditional parameterization method. The 3D case studies were three geological scenarios: binary and bimodal channelized systems, and a three-facies channel-levee-mud system. The algorithm was successfully applied for history matching with ESMDA for the bimodal channelized system showing uncertainty reduction, although it requires improvements.

Bao et al. (2022) compared the performance of VAE and GAN in flow and transport DA using ESMDA. Four cases with categorical variables and three cases with continuous variables were conducted to test the performance of coupling ESMDA with deep learning models. The results showed that VAE was more successful in terms of DA performance and GAN was more efficient for reconstructing channel structure with similar properties as in the training image. The authors highlighted that VAE models don't need adversarial training like GANs, that is time-consuming and may be unstable.

Ling and Jafarpour (2024) propose the Style-based Generative Adversarial Networks in the second version (StyleGAN2) for parameterization of complex subsurface flow properties and subsequent ESMDA application comparing the results with CVAE and GAN. This architecture was evaluated it using groundwater pumping tests and two-phase fluid flow experiments. The results showed that parameterization with StyleGAN2 provide superior performance in terms of reconstruction fidelity and flexibility. The authors highlighted that StyleGAN2 have direct implications for representing complex subsurface heterogeneity using low-dimensional latent variables, including improved fidelity and control of image attributes. They also highlighted a notable implication for model calibration was the increased regularity of the latent space (z-space), which implies that two points that are close in the latent space also remain close in the original spatial domain. After that, Ahn and Choe (2025) propose a novel method that enhances history matching in reservoir simulations by integrating a geological-style-mixing approach with GAN-based optimization using StyleGAN3, a framework capable of producing a variety of geological styles. Thus, the focus of this work was on generating diverse geological models to expand the initial ensemble to be used in DA, not using StyleGAN3 in the parameterization process. The proposed method was compared with the previously developed convolutional neural network-principal component analysis (CNN-PCA) and demonstrated similar history matching performance.

Federico and Durlofsky (2025) used a variational autoencoder for dimension reduction and a U-Net for the denoising process for latent diffusion model. This model was applied in 2D three-facies (channel-levee-mud) systems obtained visually consistent realizations with samples used and significant uncertainty reduction in the DA process. Although good quality images were obtained, the production and injection matching results were limited to reducing uncertainties in both cases, indicating a need for improved matching.

Our previous work stemmed from the findings by Canchumuni et al. (2021) and Bao et al. (2022) that GANs are capable of generating high-quality images, albeit with poor matching, whereas VAEs exhibit the opposite behavior. Consequently, we proposed applying the hybrid VAE-GAN method to leverage the advantages of both. Indeed, the VAE-GAN model achieved output quality comparable to that of GANs, alongside assimilation results similar to those of VAEs (Sampaio et al., 2026). We evaluated the results across both categorical and continuous test cases using standard metrics. We also highlighted that VAE-GAN can be time-consuming and unstable, as it requires adversarial training between three networks.

In the current work, inspired by the observations of Karras et al. (2019) regarding the StyleGAN model, specifically that the $w$-space is much more linear and disentangled than the highly entangled the $z$-space, we introduce a new parameterization method for non-Gaussian reservoir properties in ensemble-based DA, utilizing the intermediate latent space of second generation of StyleGAN. To demonstrate the effectiveness of this method, we compared our approach with DA using the conventional latent $z$-space, as well as against state-of-the art models in these problems, such as Latent Diffusion models (Federico and Durlofsky, 2025) and VAE-GAN (Sampaio et al., 2026).

## 2. METHODOLOGY

In this section, we present the methodology developed in this work. It is divided into two main parts. The first part involves modeling and training three types of deep learning models to compare their respective advantages and disadvantages: VAE-GAN, Latent Diffusion and StyleGAN2. The second part consists in making the integration with DA, using the ESMDA to perform history matching. All models used the latent $z$-space for DA, except for StyleGAN2 model, which is also evaluated using its intermediate latent $w$-space. The images (realizations) in this study consist of 2D permeability maps defined on a 48 x 48 Cartesian grid.

## 2.1. Deep Learning Models

In this first part, we briefly describe the three deep learning models used in this work.

### *2.1.1. Variational Autoencoder Generative Adversarial Network (VAE-GAN)*

The Variational Autoencoder Generative Adversarial Network (VAE-GAN) combines the advantages of VAE and GAN to improve image generation quality while preserving a structured latent representation. The VAE encoder learns a latent representation of the image and imposes a probabilistic distribution on this latent space ($z$-space), while the VAE decoder reconstructs images from the latent representation. Additionally, the GAN discriminator distinguishes real images from images generated by the decoder. In this architecture, the VAE is responsible for ensuring that the latent space is well structured and regularized, whereas the GAN improves the quality of generated images by forcing the decoder to generate more realistic images. The objective function combines the VAE and GAN losses to balance reconstruction and geological realism (Sampaio et al., 2026).

### *2.1.2. Latent Diffusion Model (LDM)*

The genesis of Latent Diffusion Models (LDMs) lies at the convergence of two distinct lines of research in deep generative learning: Denoising Diffusion Probabilistic Models (DDPMs) and Autoencoders (Rombach et al., 2022). The primary motivation for this fusion was to address the core limitation of original DDPMs: the high computational cost of training and inference, inherent to operating directly in high-dimensional pixel space (Ho et al., 2020). The resulting framework is composed of two stage architecture: an autoencoder and a diffusion model. In the autoencoder stage, an encoder compresses input images into a lower-dimensional latent representation ($z$-space), capturing essential spatial features while discarding unnecessary details. Conversely, the decoder reconstructs images from these latent codes, ensuring that the latent space preserves meaningful information. In the diffusion process, the model learns to generate new latent representations by gradually denoising random noise. During the training step, referred to as the forward process, Gaussian noise is progressively added to clean latent codes across multiple time steps. In this model, a U-Net model learns to predict the noise added at each step, trained with a simple MSE loss between predicted and actual noise. After that, in the generation part, called as reverse process, it starts from pure random noise in the latent space and the trained U-Net iteratively removes noise step by step. In each step predicts and subtracts the noise component, gradually revealing a structured latent code. The clean latent code produced by the reverse diffusion is passed through the decoder to generate the final image.

The LDM used in this work consists of a variational autoencoder (VAE) and a U-Net, similar to the architectures employed by Ronneberger et al. (2015) and Federico and Durlofsky (2025).

#### *2.1.3. Style-Based Generative Adversarial Network (StyleGAN)*

The Style-based Generative Adversarial Networks (StyleGAN) architecture emerged from a progressive evolution of generative adversarial networks, driven by the need for greater control over the image synthesis process and higher output quality (Karras et al., 2018, 2019, 2020a, 2020b, 2021). The foundational GAN framework was introduced by Goodfellow et al. (2014), establishing the adversarial training paradigm, where a generator and discriminator compete in a min-max game. However, early GANs suffered from training instability and mode collapse, issues that were partially addressed by architectural innovations. Furthermore, manipulating the latent code to obtain a desired change in the generated image is challenging, since the mapping between the latent and image spaces is highly nonlinear. Consequently, changes to a latent code may simultaneously affect multiple semantic attributes of the generated image, a property usually referred to as entanglement. In this regard, Karras et al. (2019) demonstrated significant improvements in image quality and control over the disentanglement of semantic attributes with StyleGAN model. In the StyleGAN2, a random latent vector (z-space) drawn from a standard normal distribution is first transformed by a mapping network into an intermediate latent representation ($w$-space). This transformed representation is then injected at multiple resolution levels of a synthesis network via adaptive instance normalization (AdaIN) operations, which modulate the feature maps by scaling and shifting their normalized activations. This mechanism allows coarse styles to control high-level geological structures, while finer styles govern local heterogeneity. A central aspect of this architecture is the enforcement of Gaussianity within the latent space. The model imposes a statistical regularization that constrains the latent vectors to maintain zero mean, unit variance, and negligible skewness and kurtosis. This is achieved through an additional loss term that penalizes deviations from standard normal moments. As a result, the latent manifold remains smooth and continuous, enabling meaningful interpolation between generated realizations and robust sampling for downstream applications. Overall, the model operates by learning a constrained, smooth mapping from a well-behaved Gaussian latent space to realistic geological images, guided by adversarial feedback, perceptual feature matching, and explicit statistical regularization. The improvements in latent space regularity and predictability, as well as generation of spatial domain images with higher quality and robustness, are among the reasons to use StyleGAN for subsurface flow modeling, and in particular in parameterization of model calibration problems (Ling and Jafarpour, 2024)

### 2.2. Ensemble Smoother with Multiple Data Assimilation (ESMDA)

Emerick and Reynolds (2013) demonstrated that Ensemble Smoother (ES) is equivalent to a single full Gauss-Newton update step. Therefore, to handle with nonlinear forward operators, ESMDA assimilates all the observed data multiple times using an inflated measurement error covariance matrix, relating each iteration with a smaller Gauss-Newton update step. Neglecting model errors, the perfect forward model is expressed as $\mathbf{d} = g(\mathbf{m})$, where the non-linear forward operator $g(\cdot)$ (in our case, the reservoir simulator) maps the model parameters vector $\mathbf{m} \in \Re^{N_m}$ to the predicted data vector $\mathbf{d} \in \Re^{N_d}$. Here, $N_m$ and $N_d$ are the number of model parameters and observed data points, respectively. The history matching inverse problem aims to estimate the model parameter vector $\mathbf{m}$ that best reproduce a set of observed measurements ($\mathbf{d}_{\text{obs}}$). The observed data are considered a noisy realization of the true reservoir response ($\mathbf{d}_{\text{true}}$), defined as $\mathbf{d}_{\text{obs}} = \mathbf{d}_{\text{true}} + \epsilon$. The measurement error $\epsilon$ is usually drawn from a zero-mean Gaussian distribution, $\epsilon \sim \mathcal{N}(0, \mathbf{C}_{\text{D}})$, with $\mathbf{C}_{\text{D}} \in \Re^{N_d \times N_d}$ representing the covariance matrix of measurement errors. In the ESMDA analysis step, each member $j$ of model parameters vector ensemble is updated using the following equation:

$$\mathbf{m}_j^{i+1} = \mathbf{m}_j^i + \mathbf{C}_{MD}^i \left( \mathbf{C}_{DD}^i + \alpha_i \mathbf{C}_{\text{D}} \right)^{-1} \left( \mathbf{d}_{obs,j}^i - \mathbf{d}_j^i \right) \quad (1)$$

for $j = (1, \ldots, N_e)$, where $N_e$ denotes the number of ensemble members (ensemble size), $\mathbf{C}_{MD}^i \in \Re^{N_m \times N_d}$ is the cross-covariance between the model parameters and predicted data, $\mathbf{C}_{DD}^i \in \Re^{N_d \times N_d}$ is the auto-covariance of the predicted data, $\alpha_i$ is the inflation factor that damp the iteration $i$. The forward step is where each ensemble member $\mathbf{d}_j^i$ is estimated using the forward operator $\mathbf{d}_j^i = g(\mathbf{m}_j^i)$ for $j = (1, \ldots, N_e)$.

## 2.3. Geostatistical Metrics

In this section, we introduce the geostatistical metrics used to evaluate the images generation in terms of geological realism and quality.

### *2.3.1. Variogram Mean Squared Error*

The variogram measures spatial continuity by quantifying how data values differ as a function of distance. For each lag distance h, the 2D experimental variogram is computed as:

$$\gamma(h) = \frac{1}{2N(h)} \sum_{i,j} \left[ Z(x_i) - Z(x_j) \right]^2 \quad (2)$$

where $Z(x)$ is the value at location x, and N(h) is the number of point pairs separated by distance $h$ in both horizontal and vertical directions. The MSE between real and generated variograms is then calculated as the mean squared difference over all lag distances.

### *2.3.2. Connectivity Function MSE*

The connectivity function quantifies how well a generated spatial model reproduces the connectivity patterns of real data. The connectivity function, denoted as τ(h), measures the probability that two points separated by a lag distance $h$ are not only above a threshold $t$ but also belong to the same connected cluster. The MSE then computes the average squared difference between the connectivity functions of the real dataset and the generated (simulated) dataset across all lag distances. The connectivity function is defined as:

$$\tau(h) = \frac{\sum_{i,j} \mathbf{1}[Z(x_i) > t \wedge Z(x_j) > t \wedge \text{connected at } h]}{\sum_{i,j} 1[Z(x_i) > t]} \tag{3}$$

where **1**[·] is the indicator function, and the sums run over all point pairs at lag distance $h$. The MSE is computed between the connectivity functions of real and generated data. Lower MSE values indicate that the generated data faithfully preserves the spatial continuity and clustering behavior of the reference real data.

### *2.3.3. Histogram KL Divergence*

This metric compares the full pixel value distributions using the Kullback-Leibler divergence (Kullback and Leibler, 1951). After computing normalized histograms P (real) and Q (generated) with a small epsilon added for numerical stability:

$$D_{KL}(P \,||\, Q) = \sum_i P(i) log\left(\frac{P(i)}{Q(i)}\right) \tag{4}$$

A lower value indicates that the generated distribution closely matches the real data distribution.

### *2.3.4. PCA Correlation*

Principal Component Analysis (PCA) is performed on flattened real and generated images. The correlation between corresponding principal components measures how well the global variance structure is preserved:

$$\rho_k = \frac{Cov(PC_{real,k}, PC_{fake,k})}{\sigma_{real,k} \cdot \sigma_{fake,k}} \quad (5)$$

The reported metric is the mean of these correlations across the first two components. Values closer to 1 indicate strong structural similarity.

#### *2.3.5. MDS MMD*

Multidimensional Scaling first projects both real and generated samples into a low-dimensional space. The Maximum Mean Discrepancy then quantifies the distance between these two distributions using a Gaussian kernel:

$$MMD^2 = \mathrm{E}[K(X, X')] + \mathrm{E}[K(Y, Y')] - 2\mathrm{E}[K(X, Y)] \quad (6)$$

where $K(x, y) = \exp(-\gamma \|x - y\|^2)$. The bandwidth $\gamma$ is set adaptively as the inverse of twice the squared median of pairwise distances. A lower MMD indicates the generated samples are statistically indistinguishable from real ones in the MDS embedding space.

### 2.4. History Matching Metrics

In this section, we introduce the metrics used to evaluate the DA results.

#### *2.4.1. Normalized data-mismatch objective function*

For each member $j$, the normalized data-mismatch is the difference between simulated and observed data, scaled by the measurements error covariance matrix:

$$\mathrm{O}_{\mathrm{N_{d,j}}} = \frac{1}{\mathrm{N_d}} \left(\mathbf{d}_j - \mathbf{d}_{\mathrm{obs}}\right)^{\mathrm{T}} \mathbf{C}_{\mathrm{D}}^{-1} \left(\mathbf{d}_j - \mathbf{d}_{\mathrm{obs}}\right) \quad (7)$$

and its average:

$$\overline{\mathrm{O}_{\mathrm{N_d}}} = \frac{1}{N_e} \sum_{\mathrm{j}=1}^{N_e} \mathrm{O}_{\mathrm{N_{d,j}}} \quad (8)$$

#### *2.4.2. Balanced Accuracy*

Balanced accuracy is a performance metric used in classification tasks, especially when dealing with imbalanced datasets. Balanced accuracy generalizes naturally to multi-class problems by averaging the recall (true positive rate) for each class. For $C$ classes, balanced accuracy is computed as:

$$\text{Balanced Accuracy} = \frac{1}{\mathrm{C}}\sum_{\mathrm{i}=1}^{\mathrm{C}} \frac{\text{True positives}_{\mathrm{i}}}{\text{True positives}_{\mathrm{i}}+\text{False negatives}_{\mathrm{i}}}, \text{for class i} \tag{9}$$

### *2.4.3. Root Mean Square Error (RMSE)*

An ensemble of root mean square error (RMSE) for the ensemble $= \{\mathbf{m}_j\}_{j=1}^{\mathrm{N}_e}$ , with $M$ is the number of parameters and $N_e$ is the number of members of ensemble, with respect to the true model $\mathbf{m}_{\text{true}}$ is:

$$\text{RMSE}\left(\mathbf{M}, \mathbf{m}_{\text{true}}\right) = \left\{\frac{\left\|\mathbf{m}_j - \mathbf{m}_{\text{true}}\right\|_2}{\sqrt{M}}\right\}_{j=1}^{N_e} \tag{10}$$

### *2.4.4. Spread*

The spread is defined as the average RMSE between each ensemble member and the ensemble mean:

$$\text{Spread}(\mathbf{M}) = \frac{1}{N_e}\sum_{j=1}^{N_e}\frac{\left\|\mathbf{m}_j - \bar{\mathbf{m}}\right\|_2}{\sqrt{M}} \tag{11}$$

### *2.4.5. Fréchet Inception Distance (FID) and Fréchet Reservoir Distance (FRD)*

The Fréchet Inception Distance (FID) is estimated by computing the Fréchet distance between the distributions $r$ and $g$, obtained from the Inception network coding layer (Heusel et al., 2018):

$$\text{FID}(r, g) = \left\|\mu_r - \mu_g\right\|_2^2 + Tr\left(\mathbf{C}_r + \mathbf{C}_g - 2\left(\mathbf{C}_r\mathbf{C}_g\right)^{\frac{1}{2}}\right) \tag{12}$$

where $\mu$ and $\mathbf{C}$ denote the mean and covariance of a given set of samples, while $Tr(\cdot)$ calculates the trace of a matrix. For further information about Inception network and GAN metrics, reader is directed to Borji (2018). Although widely used in computer vision, computing the FID metric to assess the quality of generated reservoir realizations poses challenges, because the nature of the ImageNet dataset is very different from the geological images used in this work. For this reason, we replaced the Inception model with a Reservoir Classifier (RC) Network for computing the Fréchet Distance as proposed by (Ranazzi et al., 2024), calling this metric as the Fréchet Reservoir Distance (FRD). For both metrics, the statistics were computed over a large batch of 10,000 generated and training samples to avoid bias. For more information on the Reservoir Classifier (RC) and Fréchet Reservoir Distance (FRD) used in this work, the reader can find all the details in (Ranazzi et al., 2024).

**2.5. Integration between Deep Learning Models and ESMDA**

In this section, we present the integration framework between the generative deep learning models and ESMDA algorithm. Thus, we use generative models to learn low-dimensional and continuous data representation of the geological realizations, while ESMDA updates the corresponding latent vectors based on the available set of observed data. This workflow is accomplished through two main steps:

Step 1: Modeling and training deep learning models to learn the distribution of the dataset. After training, the weights of the generator (VAE-GAN and StyleGAN) and decoder (LDM) are saved for use in the next step;

Step 2: Employment of ESMDA to update the latent representations based on observed data. At each iteration, the ensemble of latent vectors $z$ (or also $w$ in the case of StyleGAN2) is passed through the respective generator or decoder to produce an ensemble of permeability realizations. These realizations are then fed into the simulator to compute the ensemble of ensemble of predicted data. The ESMDA is used to update the vector $\mathbf{z}$ of all models (and also $\boldsymbol{w}$ in the StyleGAN case) and the iterative process continues until the number of data assimilation iterations is reached.

## 3. CASE STUDIES

We evaluate the methodology by considering two distinct synthetic datasets: one case consisting of categorical variables (integer values) representing three facies, and the second one considering continuous variable based on the carbonate reservoir benchmark. The datasets were normalized to a range of -1 to 1, which is necessary for using the tanh (hyperbolic tangent) activation function in the output layer of models. All prior realizations, in each dataset, were generated using the same training image of the reference model.

### 3.1. Categorical training dataset

The realizations of the first dataset were built using the open-source Stanford Geostatistical Modeling Software – SGeMS (Remy et al., 2009), by applying the MPS algorithm Single Normal Simulation Equation – SNESIM (Strébelle, 2000). Using the "Stanford V" three-facies training image from Remy et al. (2009, chap. 8), we have created 80,000 categorical realizations, with a rotation angle of 45º in relation to the training image. In the images, shown here normalized for training, they have original permeability values equal to 100, 1,000 and 9,000 mD, for the colors purple, green and yellow, respectively. Figure 1 shows some unconditional realizations obtained by SNESIM algorithm. Henceforth, this dataset referred to as the 'categorical training dataset'.

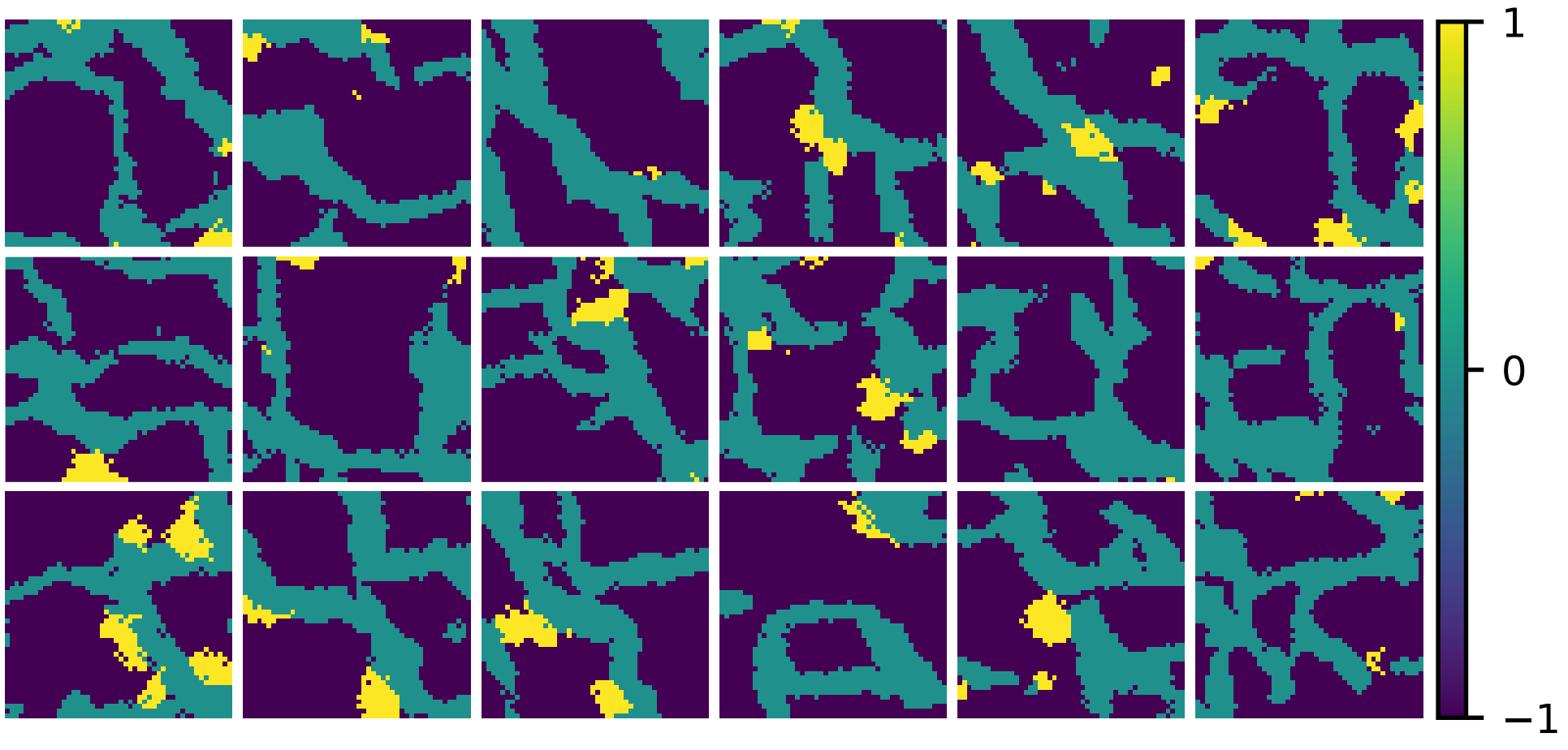

Figure 1: Random realizations of the categorical training dataset.

This case study is a 2D model that contains 48 × 48 gridblocks with the reservoir log-permeability being the only uncertain model parameter ($N_m$ = 2,304). The reference model was generated using the same approach as the training dataset. The reservoir simulation model contains 9 producers and 4 injectors with a configuration of four five-spots as we can see in Figure 2. The producers are controlled by minimum bottom-hole pressure (BHP) equal to 200 kgf/cm$^2$ and the injectors by maximum water injection rate (WIR) equal to 500 m$^3$/day. History data consists of noisy measurements at 90 days interval in 10 measurements periods ($N_d$ = 660). The covariance matrix of the measurement errors was built considering a standard deviation of 4 m$^3$/day for production rates, 3 m$^3$/day for injector rates, and 2 kgf/cm$^2$ for bottom-hole pressures.

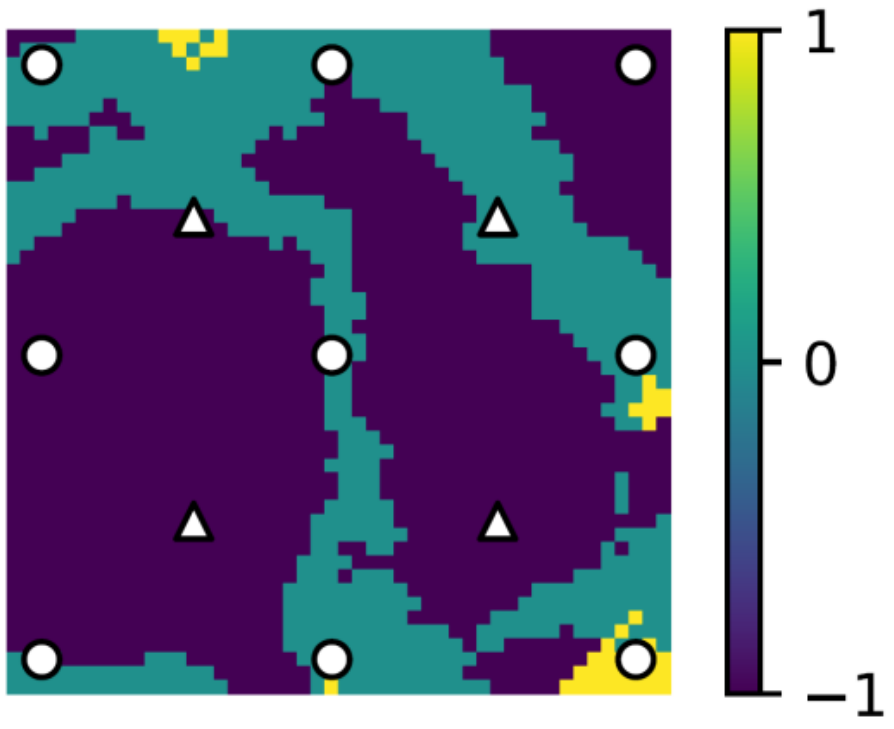

Figure 2: The true reservoir model for the first case. The locations of injectors are indicated by triangles, and those of producers by circles.

### 3.2. Continuous Training Dataset

To create the continuous training dataset in a more realistic context, we started with the UNISIM-II-H benchmark log-permeability realizations (Correia et al., 2015). The original 3D log-permeability field, which has dimensions of 46 × 69 × 30, is notably non-Gaussian due to the presence of Super-K features, thin layers with exceptionally high permeability (Meyer et al., 2000; Alqam et al., 2001). Following the approach of Ranazzi et al. (2024), we generated samples by taking multiple random 48 × 48 crops from each of the 30 vertical layers, resulting in a total of 15,000 samples (see Figure 3 for example of individual realizations). The reference model was randomly chosen from the original set of samples.

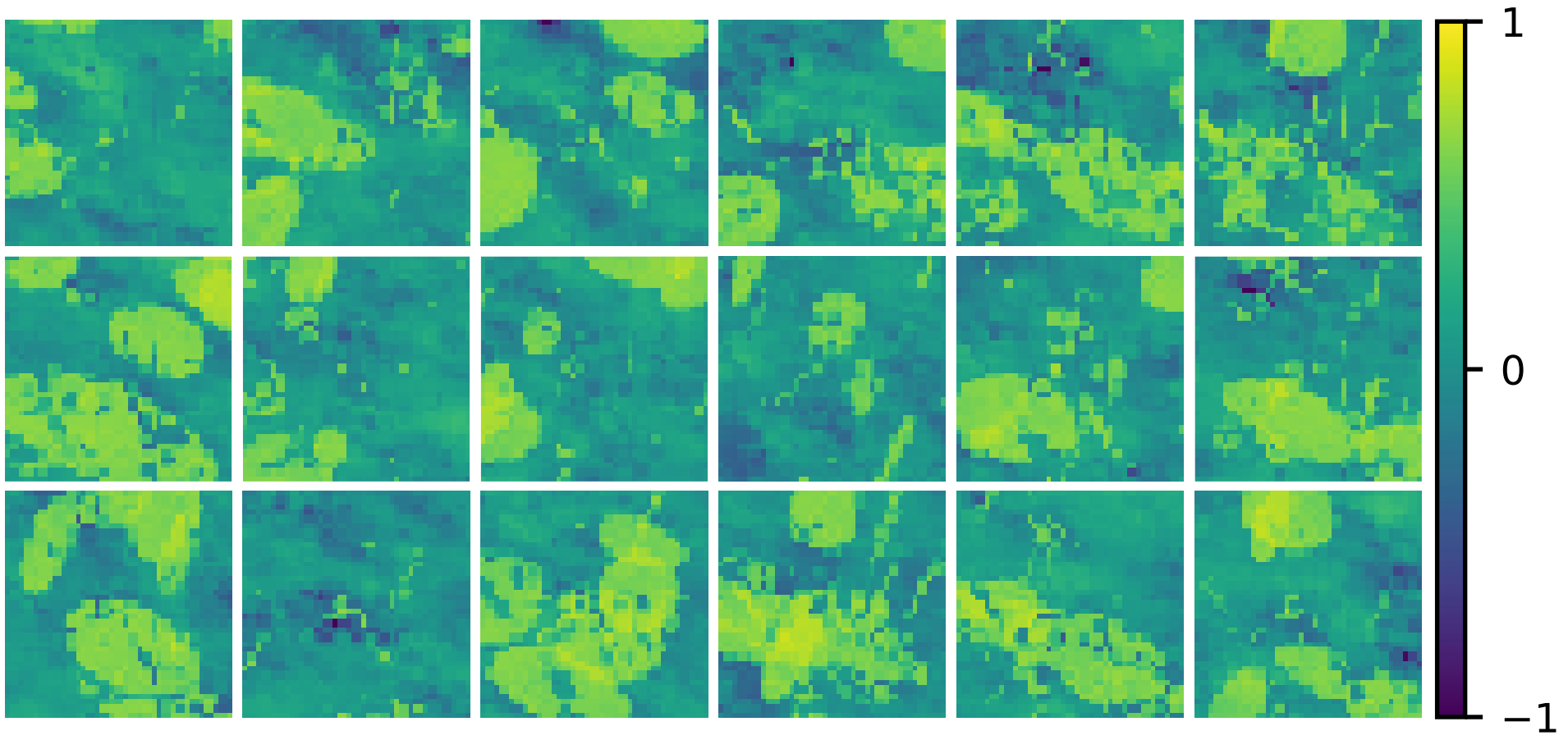

Figure 3: Random realizations of the continuous training dataset.

The reservoir simulation model contains 9 producers and 4 injectors with a configuration of four five-spots. The well control strategy is identical to that adopted in the previous case study. History data consists of noisy measurements at 90 days interval in 10 different measurements periods ($N_d = 660$). The covariance matrix of the measurement errors was built considering a standard deviation of 4 $m^3$/day for production rates, 3 $m^3$/day for injector rates, and 2 $kgf/cm^2$ for bottom-hole pressures. The true reservoir model with well positions is shown in Figure 4.

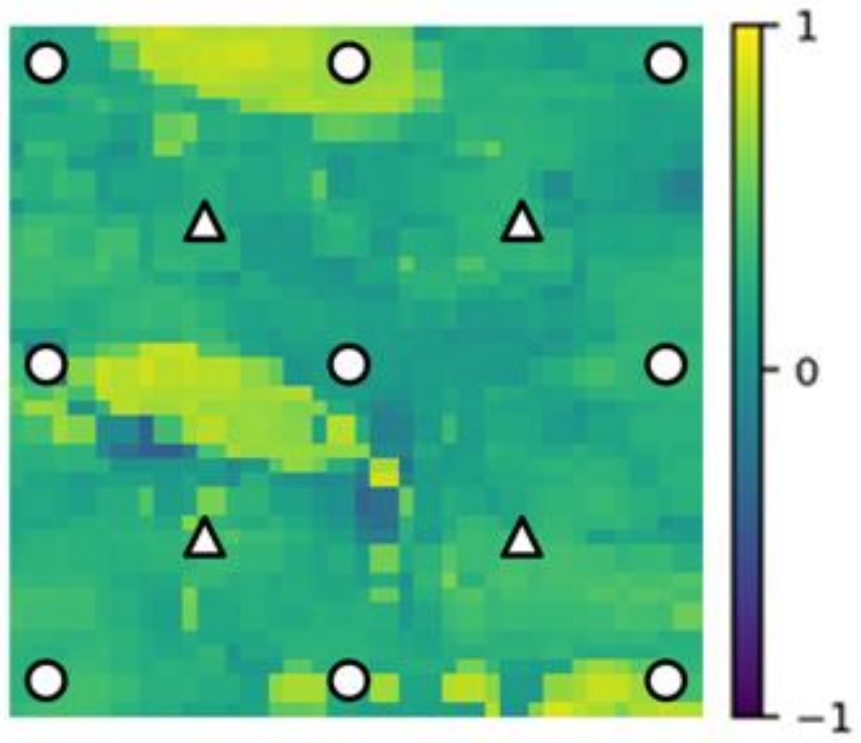

Figure 4: The true reservoir model for the second case. The locations of injectors are indicated by triangles, and those of producers by circles.

### 3.3. VAE-GAN Configuration and Structure

The VAE-GAN model was trained with a latent dimension of 512 for 150 epochs using a batch size of 32. The learning rate started at 0.0001 and followed an exponential decay schedule with a decay rate of 0.95 applied every 1,000 steps. The KL divergence loss was weighted by a beta parameter of 0.2, while the perceptual loss, computed using feature extractors based on InceptionV3 and a custom reservoir classifier, was weighted by a gamma of 0.1. The Leaky ReLU activation used an alpha slope of 0.2. The encoder consisted of three convolutional layers with 58, 116, and 230 filters, each followed by Leaky ReLU activation and batch normalization, plus a dropout rate of 0.3 before the dense layers that output the latent mean and log variance. The decoder mirrored this structure with three transposed convolutional layers using the same filter sizes in reverse order and a final tanh activation. The discriminator used three convolutional layers with the same filter progression, but incorporated spectral normalization, Gaussian noise with a standard deviation of 0.1, and dropout of 0.3. The total generator loss combined an L2 plus L1 reconstruction loss, the weighted KL divergence, the weighted perceptual loss, and an adversarial loss defined as the negative mean of the discriminator's output on fake samples. The discriminator was trained with a hinge loss using separate terms for real and fake images. Both the generator and discriminator were optimized with Adam using a $\beta_1$ of 0.5 and gradient clipping set to a maximum norm of 1.0. Early stopping with a patience of 50 epochs, monitoring the validation loss, was implemented to prevent overfitting. The Figure 5 below shows the schematic structure of the VAE-GAN model used in this work.

The VAE-GAN model contains approximately 16.7 million trainable parameters. The complete details of all models can be found in the link: [https://github.com/LASG-USP/Parameterization_StyleGAN](https://github.com/LASG-USP/Parameterization_StyleGAN).

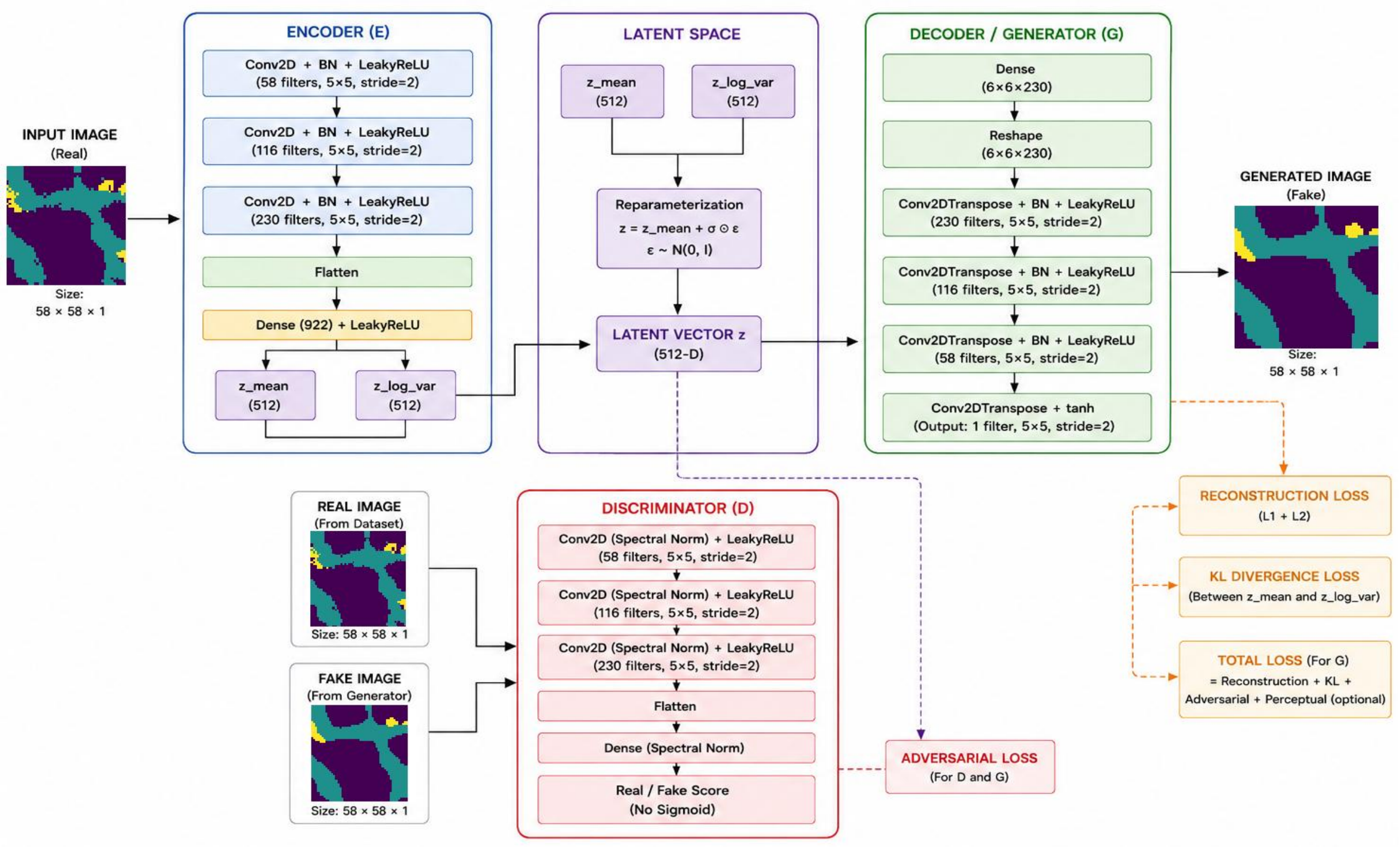

Figure 5: Schematic structure of the VAE-GAN.

### 3.4. Latent Diffusion Configuration and Structure

The encoder compresses the input image to a latent representation, where the latent dimensionality was set to 512. The encoder architecture comprises four convolutional blocks with progressively increasing filter sizes (32, 64, 128, 256), each employing 3×3 kernels with stride 2 for downsampling, followed by Leaky ReLU activation with a slope of 0.2, batch normalization, and a dropout rate of 0.4. The final convolutional features are flattened and passed through a dense layer of size 256 (with Leaky ReLU and dropout 0.3) before projection to the latent space (z-space), parameterized as a Gaussian with mean and log-variance, with latent codes are obtained via the reparameterization trick. The decoder mirrors the encoder structure through transposed convolutions (three Conv2DTranspose blocks with 128, 64, and 32 filters, each followed by Leaky ReLU 0.2 and batch normalization), reconstructing the image from the latent code. Two decoder variants were employed: (i) a discrete decoder with a softmax output activation over K = 3 facies classes, used both as the reconstruction head during training and for categorical generation via argmax over class probabilities; and (ii) a continuous decoder with a linear output activation (single channel), whose weights are not trained but derived analytically from the trained softmax decoder by projecting the final layer's class-wise outputs onto the class values $\{-1, 0, +1\}$. The continuous decoder therefore acts as a differentiable observation operator for downstream data assimilation, producing outputs naturally bounded in $[-1, 1]$.The diffusion process operates directly on the 512-dimensional latent vectors, using T = 1000 timesteps with a linear β schedule from $1\times10^{-4}$ to = 0.02 and standard DDPM forward (q-sample) and ancestral reverse (p-sample) transitions. The denoising network follows a 1D U-Net architecture conditioned on timestep embeddings. The timestep is encoded using a sinusoidal embedding of dimension 256, followed by two dense layers with Swish activation. The resulting time embedding is projected and injected into every residual block of the network. The U-Net comprises three downsampling levels with base filters of 64, yielding encoder channel widths of 64, 128, 256, and 512, followed by two residual bottleneck blocks at 512 channels and three upsampling levels with skip connections. Each residual block applies two 1D convolutions (kernel 3) with Layer Normalization and Swish activation; the time embedding is added between the two convolutions. The final layer projects back to the latent dimension (512).

Training was conducted over 150 epochs using the Adam optimizer with an initial learning rate of $10^{-4}$, exponential decay rates $\beta_1 = 0.5$ and $\beta_2 = 0.999$, and gradient clipping with a maximum norm of 1.0 applied both through the optimizer (clipnorm=1.0). A batch size of 32 was used for both training and validation. The composite loss function combined three objectives:

$$L_{total} = L_{diffusion} + \lambda_{KL}L_{KL} + \lambda_{class}L_{class} \tag{13}$$

where $L_{diffusion}$ is the mean squared error between the true and predicted noise in the latent space, $L_{KL}$ is the Kullback–Leibler divergence between the encoder posterior and the standard normal prior, and $L_{class}$ is the sparse categorical cross-entropy between the original facies map and the decoder output.

The reconstruction weight was fixed at β = 1.0. The KL weight ($L_{KL}$ ) was adaptive, initialized at $1\times10^{-4}$ and adjusted at the end of each epoch to keep $L_{KL}$ near a target of 50 (tolerance ±30%, i.e., band [35, 65]), using a ±10% multiplicative adjustment per epoch, bounded to [$1\times10^{-7}$, $5\times10^{-3}$], with a 5-epoch warmup during which the weight is held constant.

Learning-rate scheduling employed a strategy, reducing the learning rate by a factor of 0.5 if the validation loss did not improve for 5 consecutive epochs, down to a minimum of $10^{-6}$. Early stopping with a patience of 100 epochs, monitoring the validation loss and restoring the best weights, was implemented to prevent overfitting.

The Latent Diffusion model contains approximately 14.4 million trainable parameters. The Figure 6 below shows the schematic structure of the LDM used in this work.

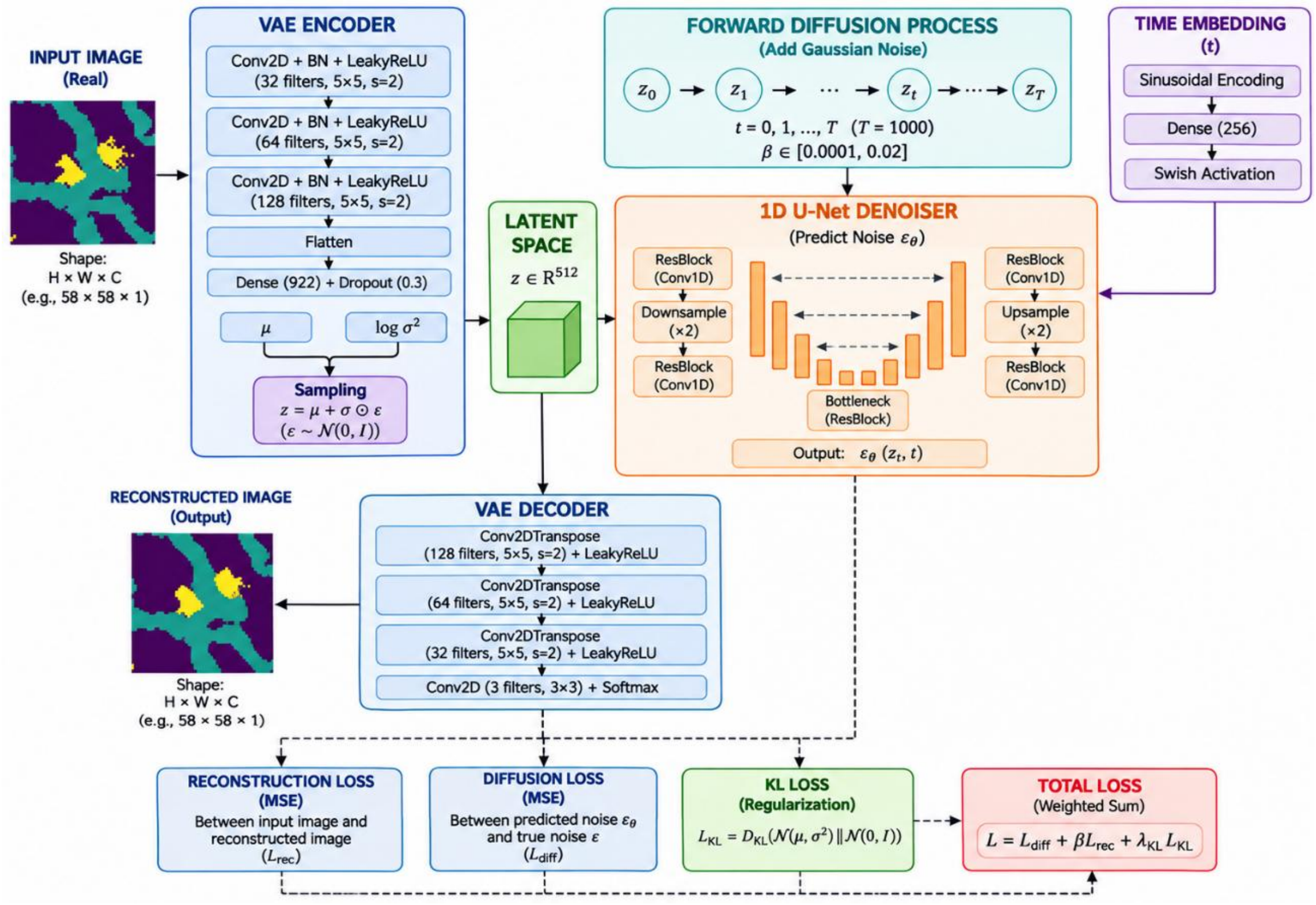

Figure 6: Schematic structure of the Latent Diffusion model.

### 3.5. StyleGAN Configuration

The generative model was trained using a Wasserstein GAN with gradient penalty (WGAN-GP) framework, with specific architectural and optimization parameters selected to ensure stable convergence and geostatistical fidelity. The generator employs a latent dimensionality of 512 (z-space), from which random vectors are drawn from a standard normal distribution. A mapping network composed of three dense layers with Leaky ReLU activations and dropout regularization transforms the latent vector (z-space) into an intermediate style representation (w-space). The synthesis network begins with a learned constant tensor of size 4×4, which is progressively upsampled through four stages using bilinear interpolation. Each upsampling stage doubles the spatial resolution, followed by convolutional layers and Gaussian-regularized adaptive instance normalization operations that inject style information. The final layer employs a convolutional operation with hyperbolic tangent activation to produce the output image. The generator applies L2 regularization with a coefficient of 0.01 throughout its layers. The discriminator processes input images through a series of five convolutional blocks with Leaky ReLU activations, dropout regularization, and average pooling operations for progressive downsampling. The architecture includes a dedicated feature extraction pathway that captures intermediate representations used for perceptual loss computation. The final classification head consists of fully connected layers that produce a scalar critic score. L2 regularization is similarly applied to all discriminator layers. Training was performed using the Adam optimizer with distinct learning rates: 0.0001 for the generator and 0.0004 for the discriminator, following the convention of training the discriminator more aggressively. The Adam momentum parameters are set to $\beta_1 = 0.5$ and $\beta_2 = 0.9$. The batch size is configured to 32 samples. Training proceeds for 150 epochs with the batch order shuffled at each epoch. Label smoothing was applied with a factor of 0.1 to prevent discriminator overconfidence. Gradient clipping is enforced with a maximum norm of 0.5 for both generator and discriminator gradients. Gaussian noise with standard deviation of 0.05 is added to both real and generated images before discriminator evaluation, improving robustness. Latent Gaussian regularization is weighted by a factor of 0.1 in the generator loss, penalizing deviations from zero mean, unit variance, zero skewness, and zero excess kurtosis in the latent vectors. Feature matching loss from intermediate discriminator representations is incorporated with a weight of 0.1. Learning rate decay of 0.9 was applied when the discriminator loss fails to improve for five consecutive epochs. An adaptive scheduler may increase learning rates toward their initial values after ten epochs of stable training, where discriminator balance (real and fake outputs summing near zero) is maintained. Label smoothing was increased up to 0.3 in response to persistent discriminator overconfidence, and gradient clipping is halved when gradient norms exceed 100. We also applied early stopping with a patience of 50 to prevent overfitting.

The StyleGAN2 model contains approximately 14.9 million trainable parameters. The Figure 7 below shows the schematic structure of the StyleGAN2 model used in this work.

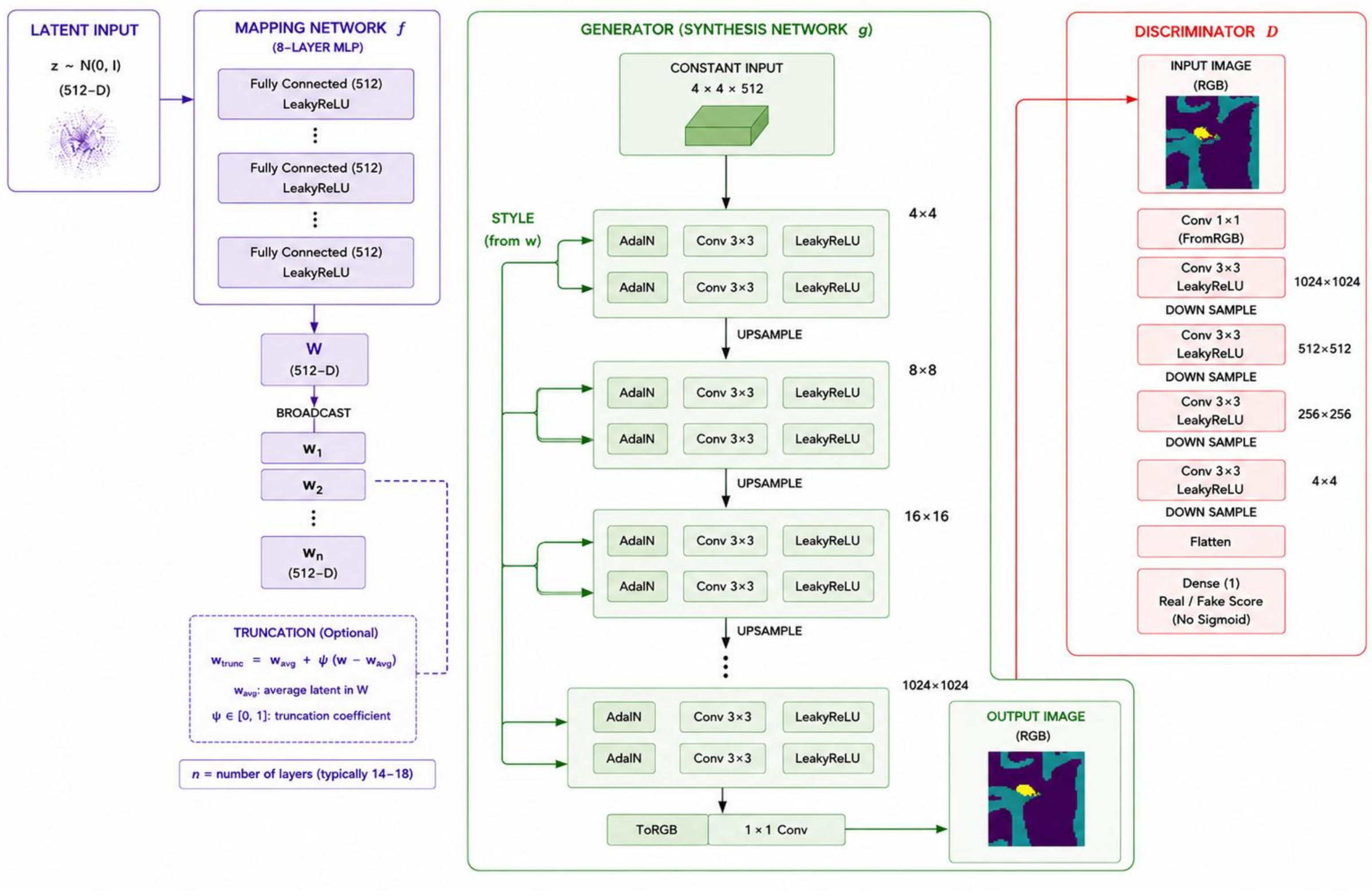

Figure 7: Schematic structure of the StyleGAN2 model.

## 4. RESULTS AND DISCUSSIONS

For each case study, we present the results of training of the models followed by the results of DA. All training process was performed in the same standard Nvidia GPU (GeForce RTX 3060 with 16 GB of memory) of a stand-alone computer, using the machine learning framework TensorFlow 2.10 (Abadi et al., 2015). In the DA step, we used 512 ensemble members and 32 iterations for ESMDA. In this way, the latent dimensions size was 512 for all models and cases.

### 4.1. Case Study 1: Categorical training dataset

#### *4.1.1. Training of Models*

***Variational Autoencoder Generative Adversarial Network (VAE-GAN)***

The VAE-GAN model was trained until the early stopping criterion was reached, reaching the FID and FRD, estimated using 10,000 samples, after the end of iterations, respectively equal to 299.7 and 0.17. The training time was 7 hours and 6 minutes, showing that the computational cost of training was the highest of all models, as it involves three models in the adversarial training. It is also the model with the highest number of trainable parameters. We can see in Figure 8 the high-quality of the images reconstructed by the decoder's model.

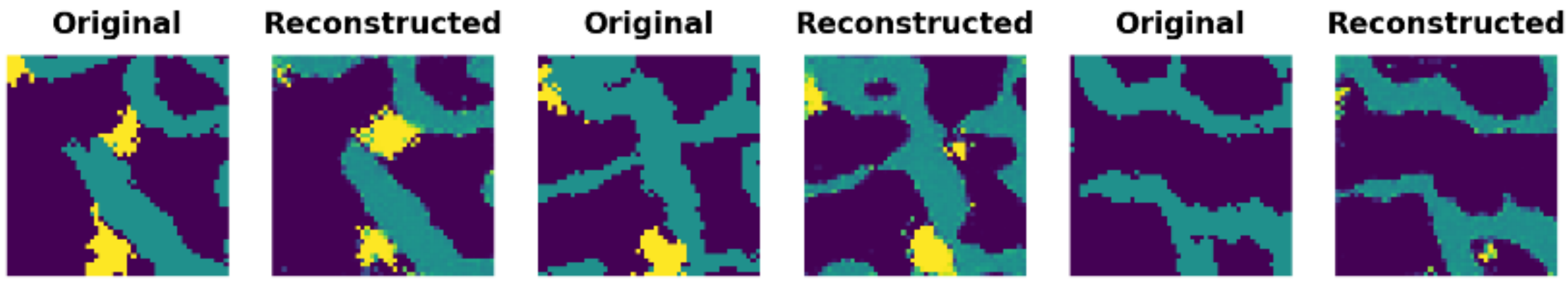

Figure 8: Three examples of random images from the set along with the reconstructed images by VAE-GAN for case study 1.

Figure 9 shows the total error for the training and the validation sets. This graph is the combination of all losses of this model: reconstruction, KL, generator, discriminator and perceptual losses. The curves show stabilization in errors after 60 epochs. We can observe initial instability until stability is reached, since it is necessary to train three networks at the same time.

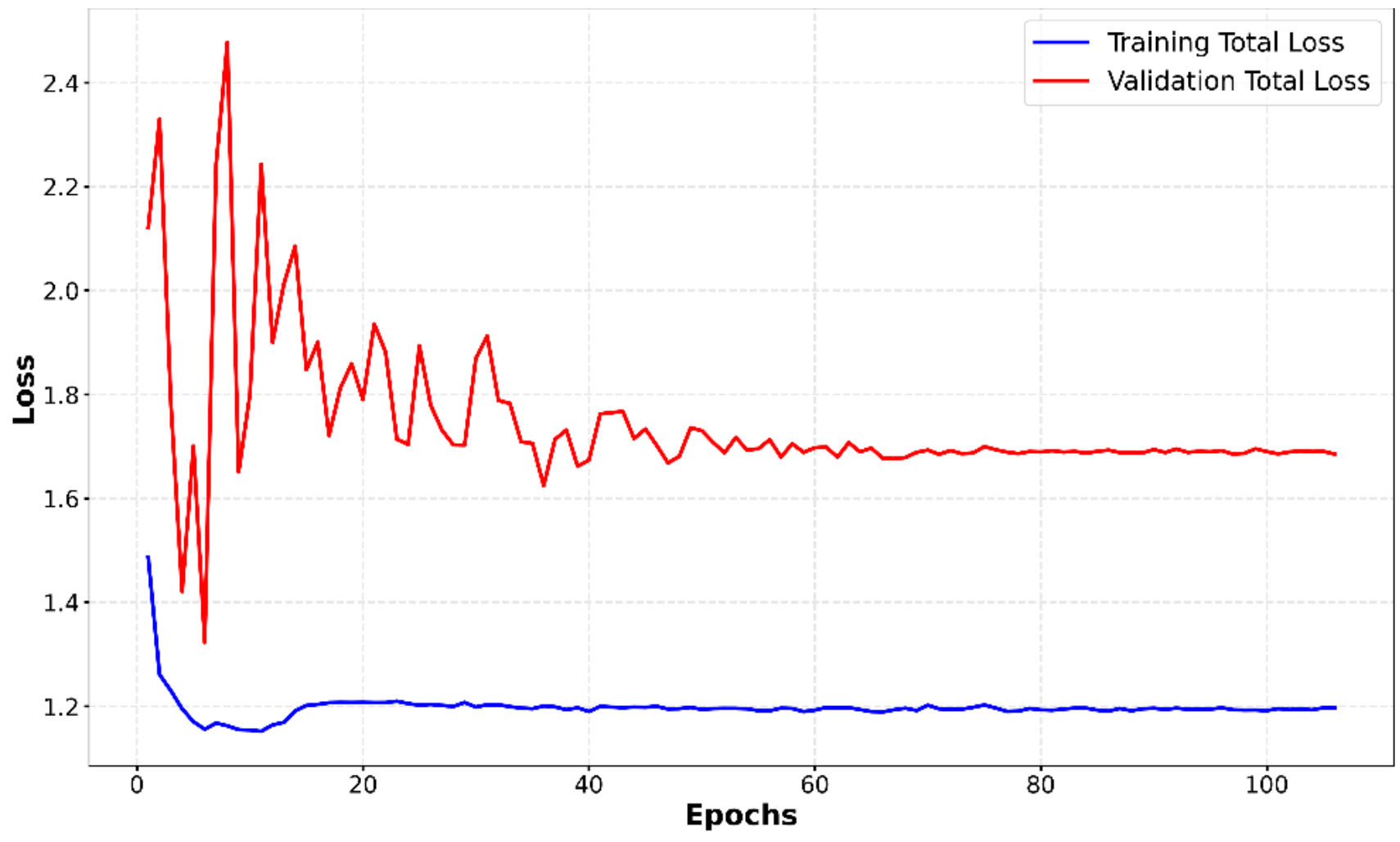

Figure 9: Total loss versus epochs with VAE-GAN in the categorical case.

***Latent Diffusion***

We trained the Latent Diffusion model (LDM) after found the better structure and configuration using the categorical training dataset until 150 epochs when stopped in the early stopping criterion. The training time was 6 hours and 49 minutes, almost the same time of VAE-GAN model. We obtained the FID and FRD, estimated using 10,000 samples, after the end of iterations, respectively equal to 424.8 and 1.0, showing that the training was efficient. The quality of the images generated at the end of the training, as shown in Figure 10, with high quality, but with a certain softening of the contours.

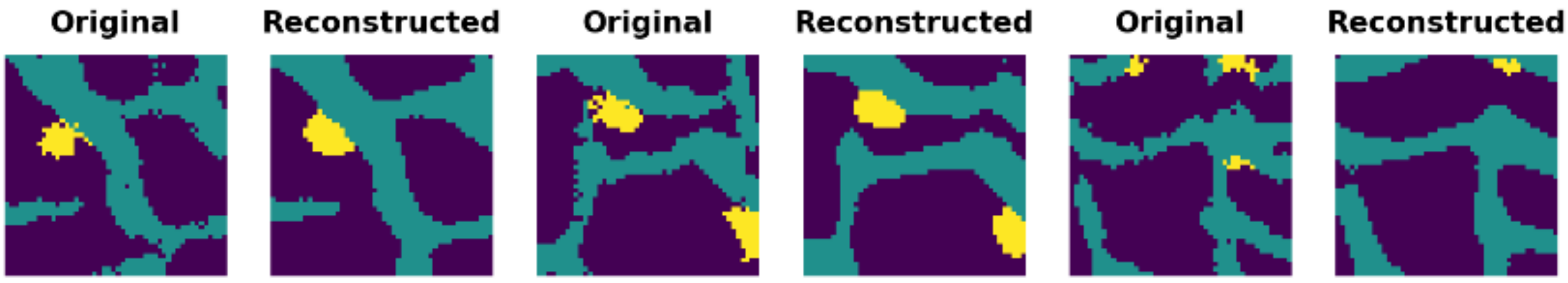

Figure 10: Images reconstructed by LDM after training with 150 epochs for case study 1.

Figure 11 show the total diffusion loss throughout the training. Thus, we can see that in this case the training was successful.

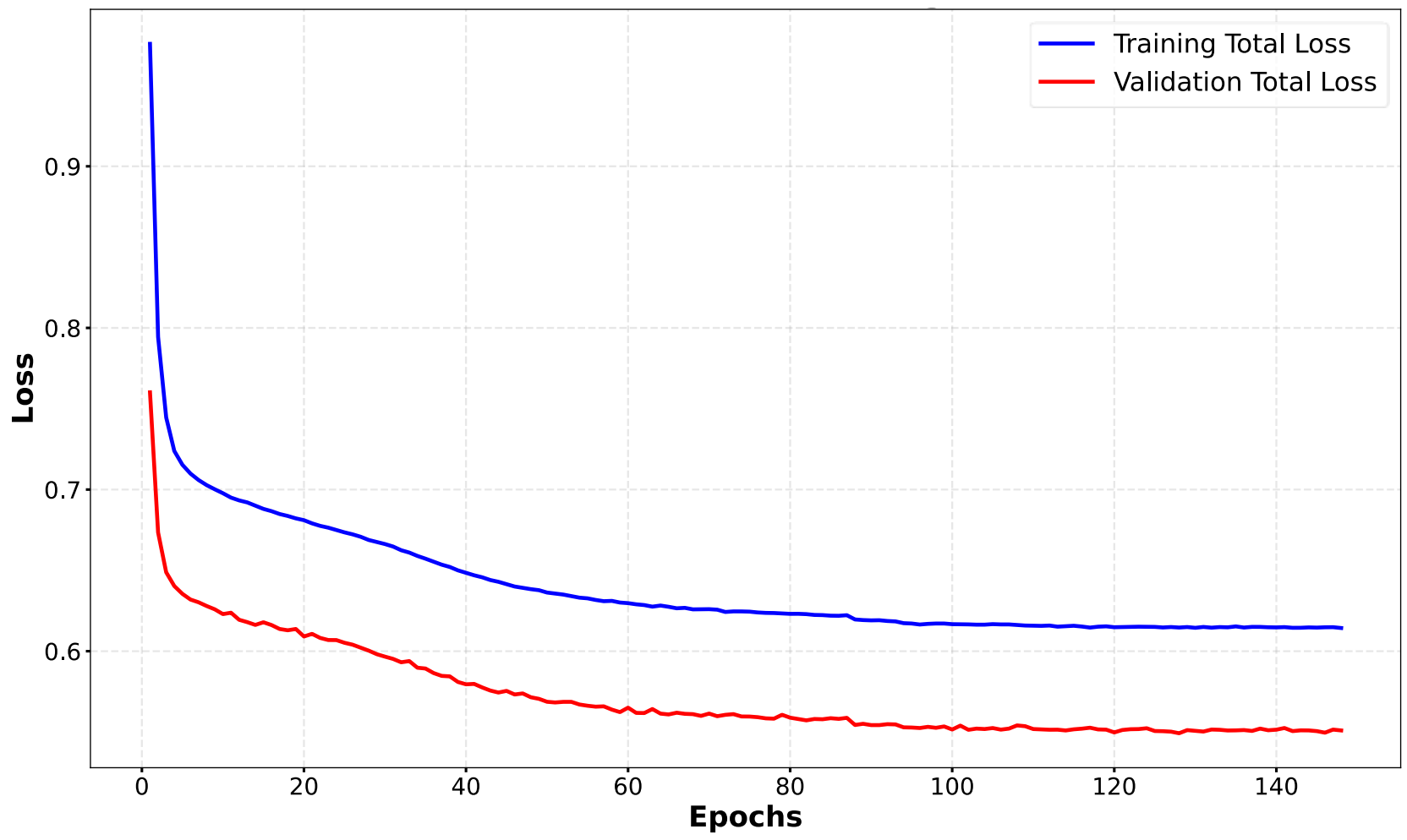

Figure 11: MSE of training and validation versus epochs with LDM across training iterations.

***Style-Based Generative Adversarial Network (StyleGAN)***

The StyleGAN model, version 2, was trained until the early stopping criterion was reached, reaching the FID and FRD, estimated using 10,000 samples, after the end of iterations, respectively equal to 300.6 and 1.28. The training time was 4 hours and 54 minutes, showing that the computational cost of training StyleGAN is less than LDM and VAE-GAN. We can see in Figure 12 the high-quality of the images generated by the generator's model.

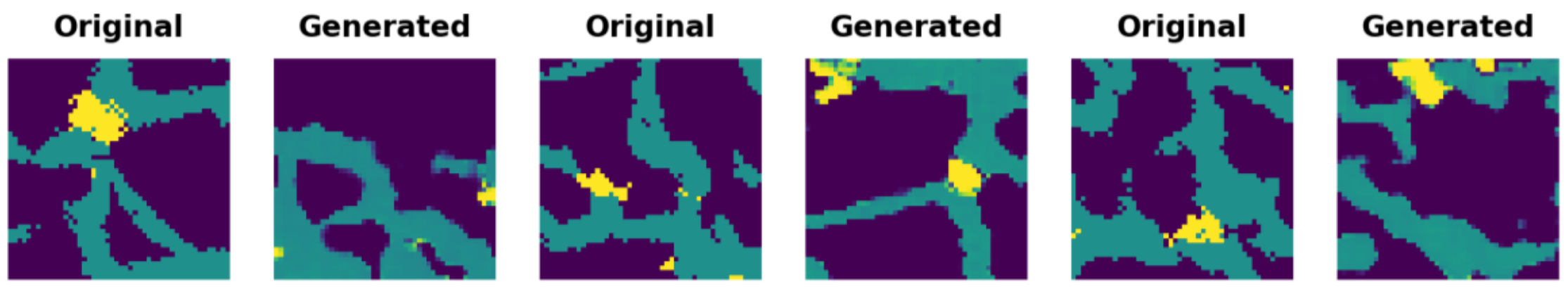

Figure 12: Three examples of random images from the set along with the generated images by StyleGAN2 for case study 1.

Figure 13 shows the decrease in the generator loss and increase in the discriminator loss curves throughout the training. We can see that both losses tend to stabilize around 50 epochs. This shows that StyleGAN2 training was efficient and sufficient.

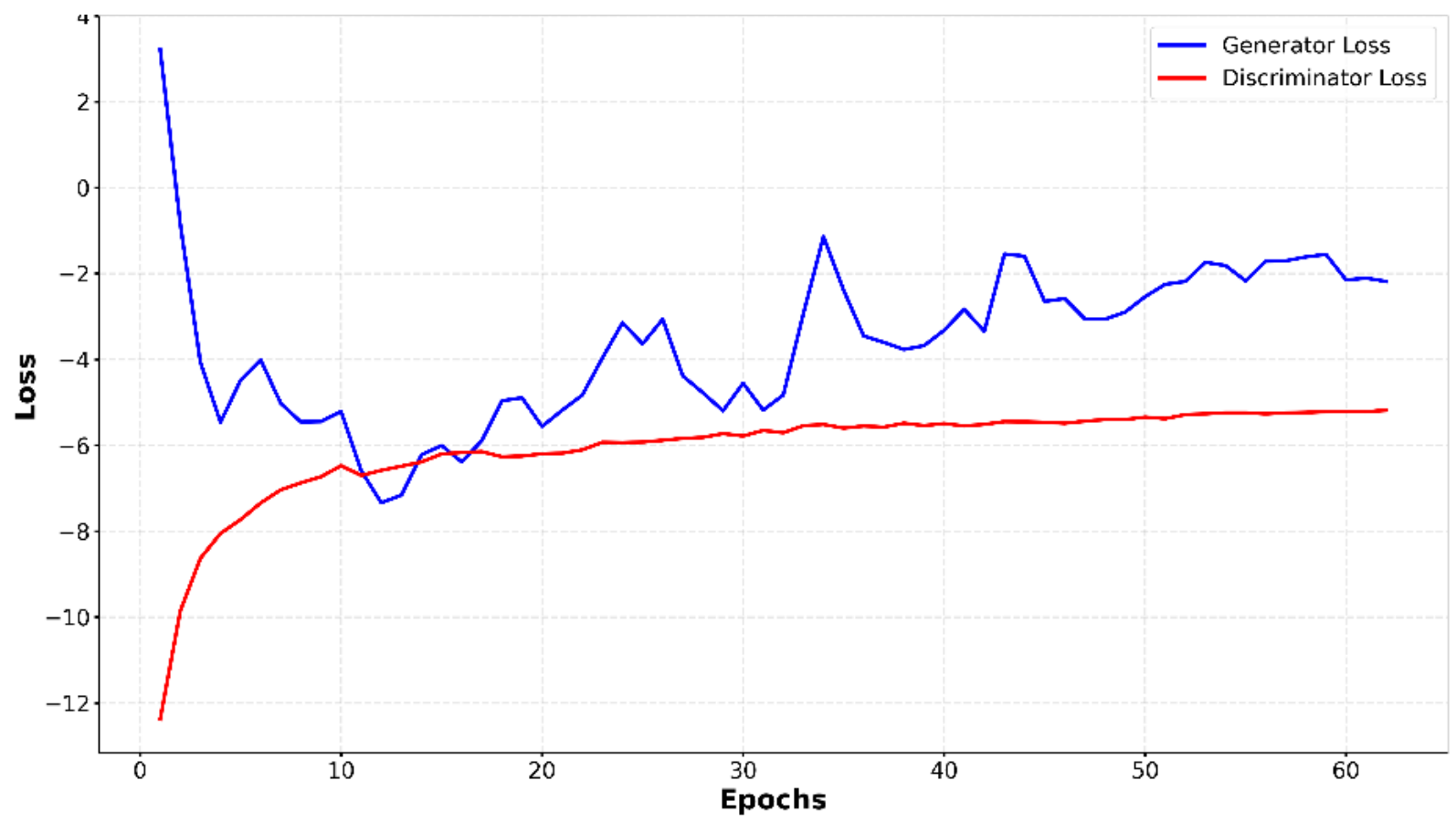

Figure 13: Graphs of generator and discriminator loss throughout the training with StyleGAN2 for case study 1.

In the Table 1, we summarize the results of the FID and FRD metrics at the end of training for the three models. We can conclude that the FRD metric was more appropriate than the FID for our case study and that the values obtained were low, demonstrating that the training of all models was efficient. It is interesting to note that the VAE-GAN reach the lowest FRD in comparison to LDM and StyleGAN, highlighting the high quality of the generated images of all models.

Table 1: Results of FID and FRD of models in the categorical case.

| | FID | FRD |
|---|---|---|
| **VAE-GAN** | 299.7 | 0.17 |
| **Latent Diffusion** | 424.8 | 1.0 |
| **StyleGAN** | 300.6 | 1.28 |

To evaluate the quality of images generated by deep learning models in geological terms, we used main geostatistical metrics, such as: variogram (MSE), connectivity (MSE), histogram KL, PCA correlation and MDS MMD (Maximum Mean Discrepancy). In the three generative models show clearly different strengths and weaknesses across the five geostatistical metrics. StyleGAN achieves the best spatial continuity, with a variogram MSE of 0.001882 and the lowest MDS MMD of 0.06334, but it performs worst on the histogram KL (1.41158), indicating poor reproduction of facies proportions. Latent Diffusion shows the opposite profile: it is best on connectivity MSE (0.00006) and histogram KL (0.02772), but its variogram MSE of 52.7970 is about four orders of magnitude larger than the other models, which is physically implausible for facies maps. VAE-GAN is consistently intermediate, with a variogram MSE of 0.00691, connectivity MSE of 0.01480, histogram KL of 0.86318, and MDS MMD of 0.10180, showing no clear advantage on any metric. PCA correlation is near zero for all three models (0.06525, 0.03739, and 0.00716 for VAE-GAN, Latent Diffusion, and StyleGAN, respectively). Overall, StyleGAN best captures spatial continuity and global manifold structure, Latent Diffusion best captures facies proportions and connectivity, and VAE-GAN remains mid-pack.

Table 2, we can see the results for static geostatistical metrics for the case study 1. Based on the comparative analysis of static geostatistical metrics, the three generative models show clearly different strengths and weaknesses across the five geostatistical metrics. StyleGAN achieves the best spatial continuity, with a variogram MSE of 0.001882 and the lowest MDS MMD of 0.06334, but it performs worst on the histogram KL (1.41158), indicating poor reproduction of facies proportions. Latent Diffusion shows the opposite profile: it is best on connectivity MSE (0.00006) and histogram KL (0.02772), but its variogram MSE of 52.7970 is about four orders of magnitude larger than the other models, which is physically implausible for facies maps. VAE-GAN is consistently intermediate, with a variogram MSE of 0.00691, connectivity MSE of 0.01480, histogram KL of 0.86318, and MDS MMD of 0.10180, showing no clear advantage on any metric. PCA correlation is near zero for all three models (0.06525, 0.03739, and 0.00716 for VAE-GAN, Latent Diffusion, and StyleGAN, respectively). Overall, StyleGAN best captures spatial continuity and global manifold structure, Latent Diffusion best captures facies proportions and connectivity, and VAE-GAN remains mid-pack.

Table 2: Results for static geostatistical metrics in the categorical case.

| | **VAE-GAN** | **Latent Diffusion** | **StyleGAN** |
|---|---|---|---|
| **Variogram MSE** | 0.00691 | 52.7970 | 0.001882 |
| **Connectivity MSE** | 0.01480 | 0.00006 | 0.00008 |
| **Histogram KL** | 0.86318 | 0.02772 | 1.41158 |
| **PCA Correlation** | 0.06525 | 0.03739 | 0.00716 |
| **MDS MMD** | 0.10180 | 0.10482 | 0.06334 |

#### *4.1.2. Data Assimilation*

For ESMDA, we updated directly the natural logarithm of permeability (log-permeability). For DA, we will compare the results from all models. However, for StyleGAN2, we will include assimilation using the z-space, as usual, and assimilation in the intermediate latent space, known as $w$-space. It is important to mention that the only modification to the StyleGAN assimilations was the shift from latent space z to w, without altering any model configurations or hyperparameters, with the training being exactly the same.

In Figure 14, we can see the images of the true case, the priori and posteriori mean and standard deviation. We also can see that the final posterior means did not result in extreme values, successfully preserving the geological realism inherent to the prior ensemble. Furthermore, an analysis of the final standard deviation maps shows that the spatial variability was more preserved. As we know, in the ensemble-collapse scenario, the ensemble of models collapses to an almost deterministic solution, with posterior variances goes to near zero. We can observe that the VAE-GAN and LDM models were able to preserve the variance and mean values of the initial ensemble better than StyleGAN models for this case.

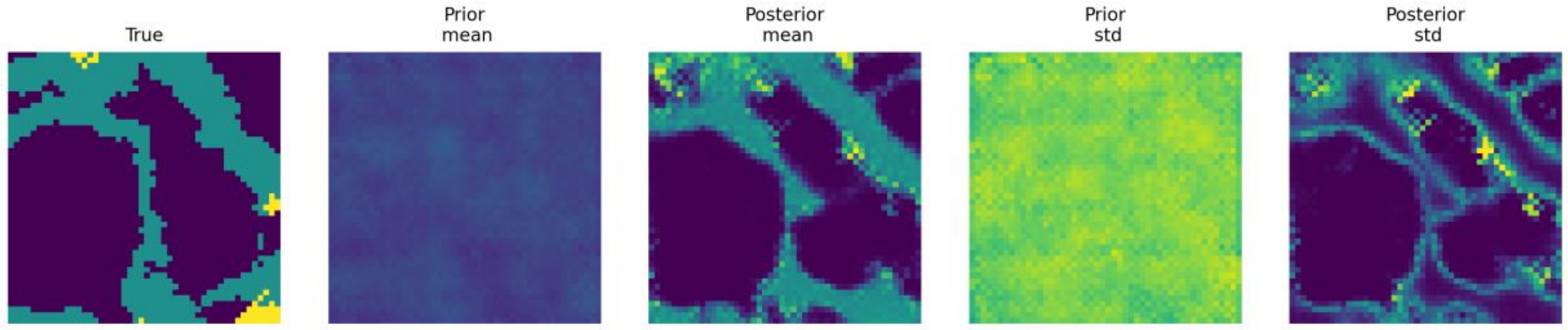

(a)

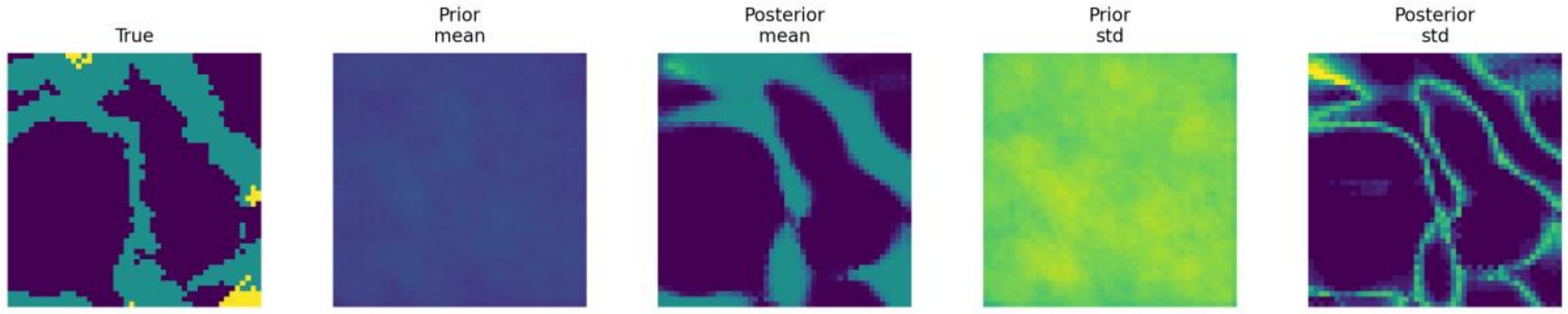

(b)

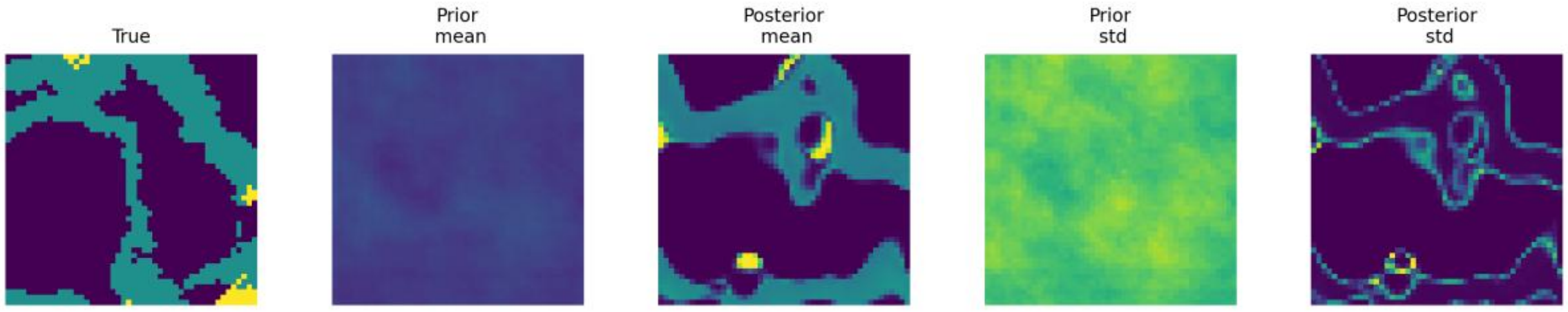

(c)

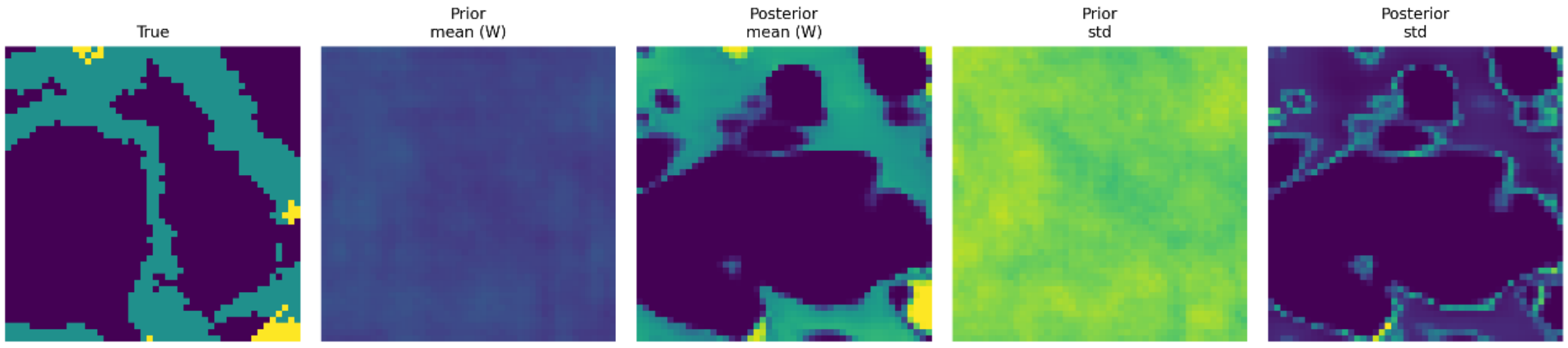

(d)

Figure 14: Images of the true case, the priori and posteriori mean, and the priori and posteriori standard deviation in the categorical case for: (a) VAE-GAN, (b) LDM, (c) StyleGAN2 (z-space) and (d) StyleGAN2 (w-space).

Analyzing the time series of production data for P1 to P6 wells in Figure 15, it is possible to confirm that in all cases resulted in reductions in terms of ensemble spread. The improvement is demonstrated visually, which depicts the oil production rate (OPR) and water production rate (WPR) time series for the producer wells. While the initial ensemble exhibits a wide spread, the updated ensemble shows a significantly narrowed spread that closely tracks the observed historical data. We can also observe that all posterior models achieved a good match for all wells, for oil and water production as a function of time. Of particular note are the excellent results achieved by the VAE-GAN and LDM models using the latent space z, and by the StyleGAN2 model using the intermediate latent space w. The results were not quite as good for the StyleGAN2 model using latent space z, likely due to the inherent entanglement characteristic of this configuration.

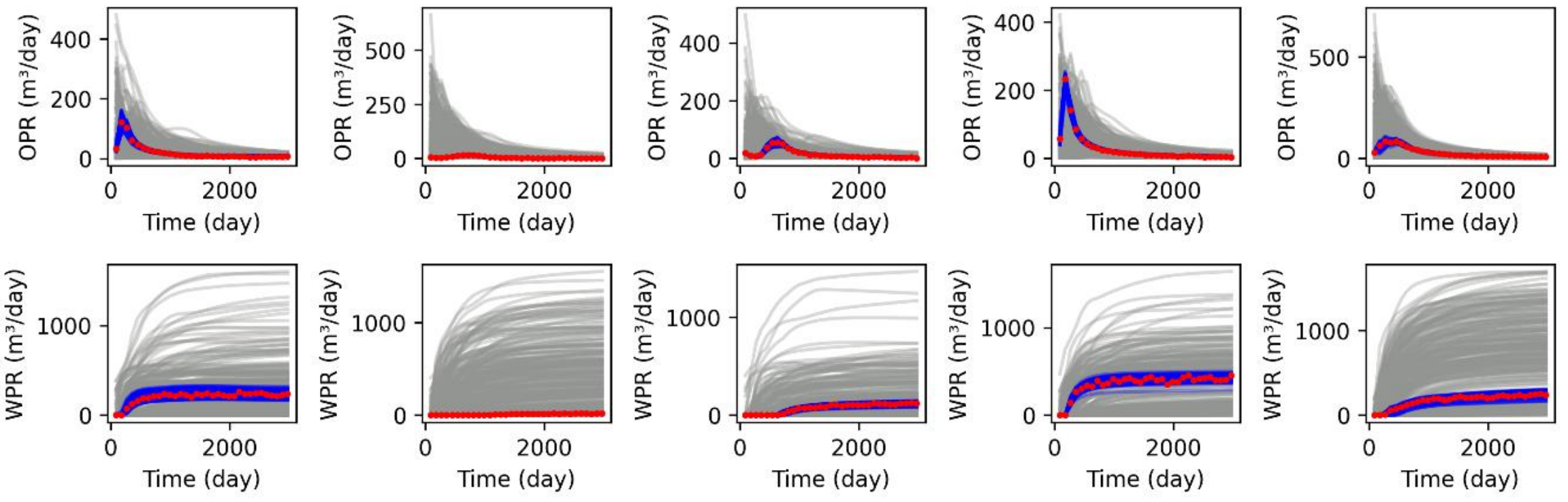
OPR (m³/day)
WPR (m³/day)
Time (day)
0
2000
200
400
250
500
1000

(a)

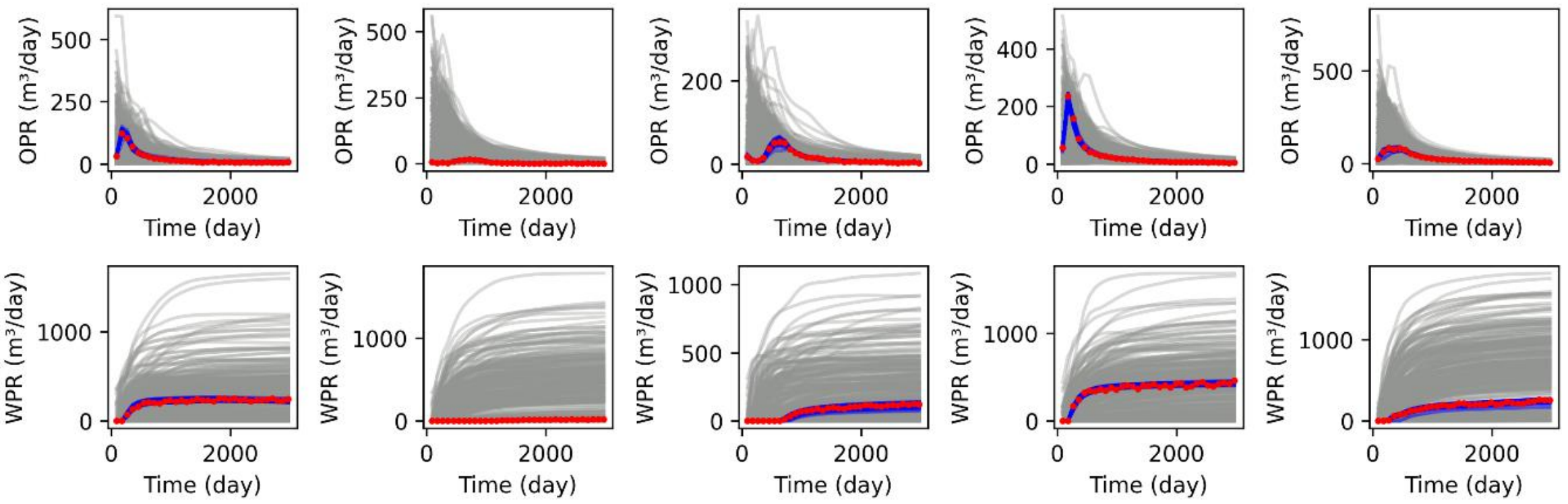
OPR (m³/day)
WPR (m³/day)
Time (day)
0
2000
200
400
250
500
1000

(b)

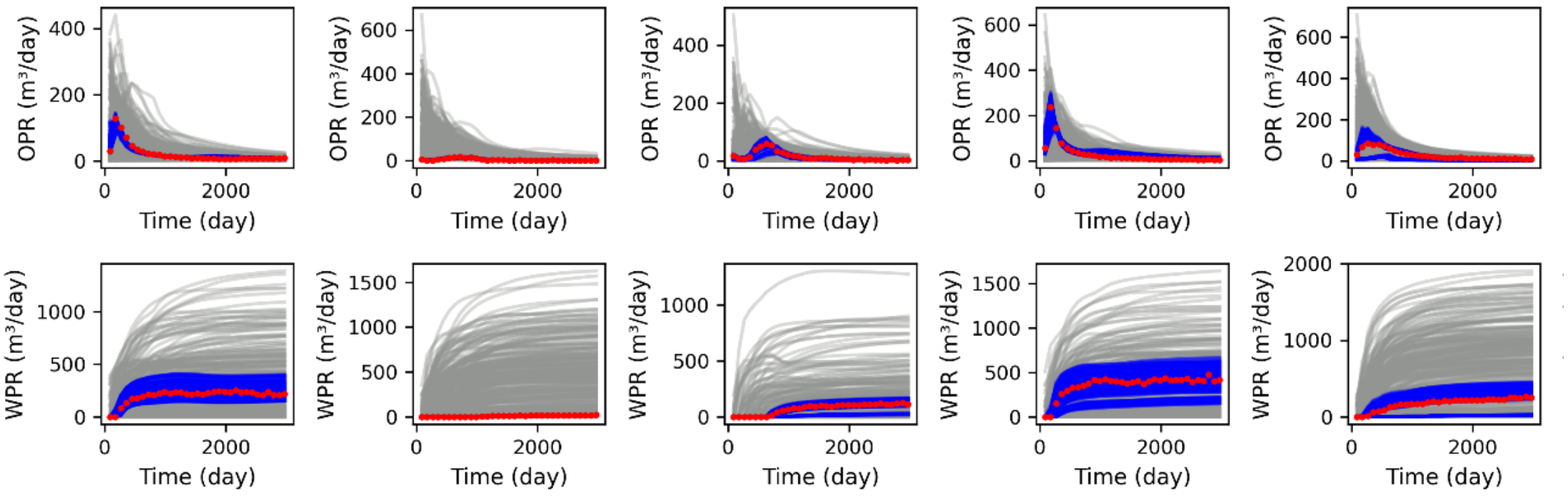
OPR (m³/day)
WPR (m³/day)
Time (day)
0
2000
200
400
600
500
1000
1500

(c)

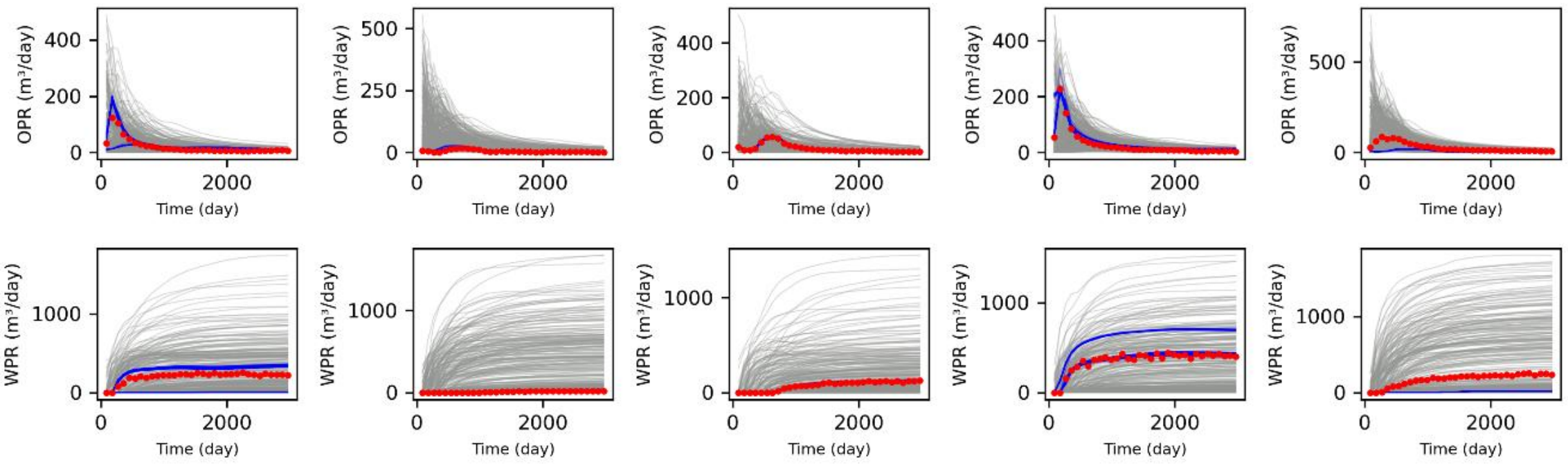

(d)

Figure 15: Time series of production data from the first five producers of oil production rate (above) and water production rate (below). Here, the gray lines represent the prior ensemble, blue lines represent the posterior ensemble, and the red dots represent the measurements in the case study 1 for: (a) VAE-GAN, (b) LDM, (c) StyleGAN2 (z-space) and (d) StyleGAN2 (w-space).

Figure 16 depicts the evolution of data mismatch, RMSE, spread and balanced accuracy across iterations. As we can see, all models achieved a satisfactory data match, as the results indicate a significant reduction in both data mismatch (DM) and RMSE, suggesting that the assimilation process effectively honored the observed data. Furthermore, LDM showed superior performance in terms of balanced accuracy compared to VAE-GAN and StyleGAN2.

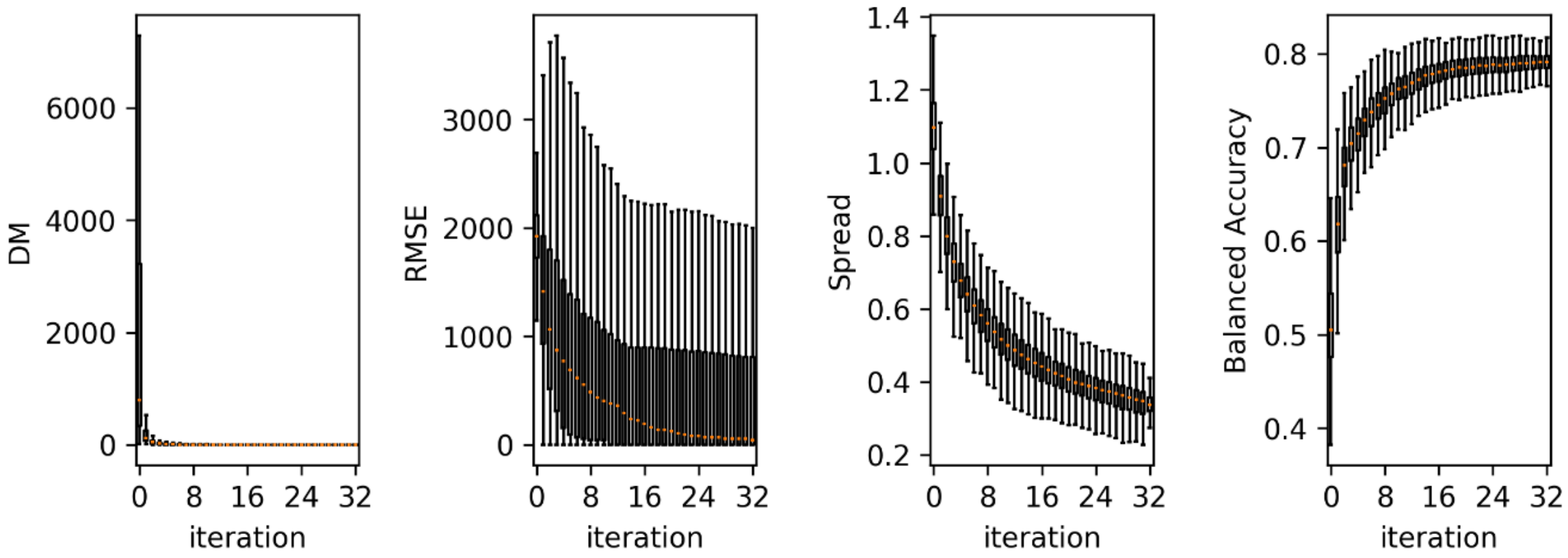

(a)

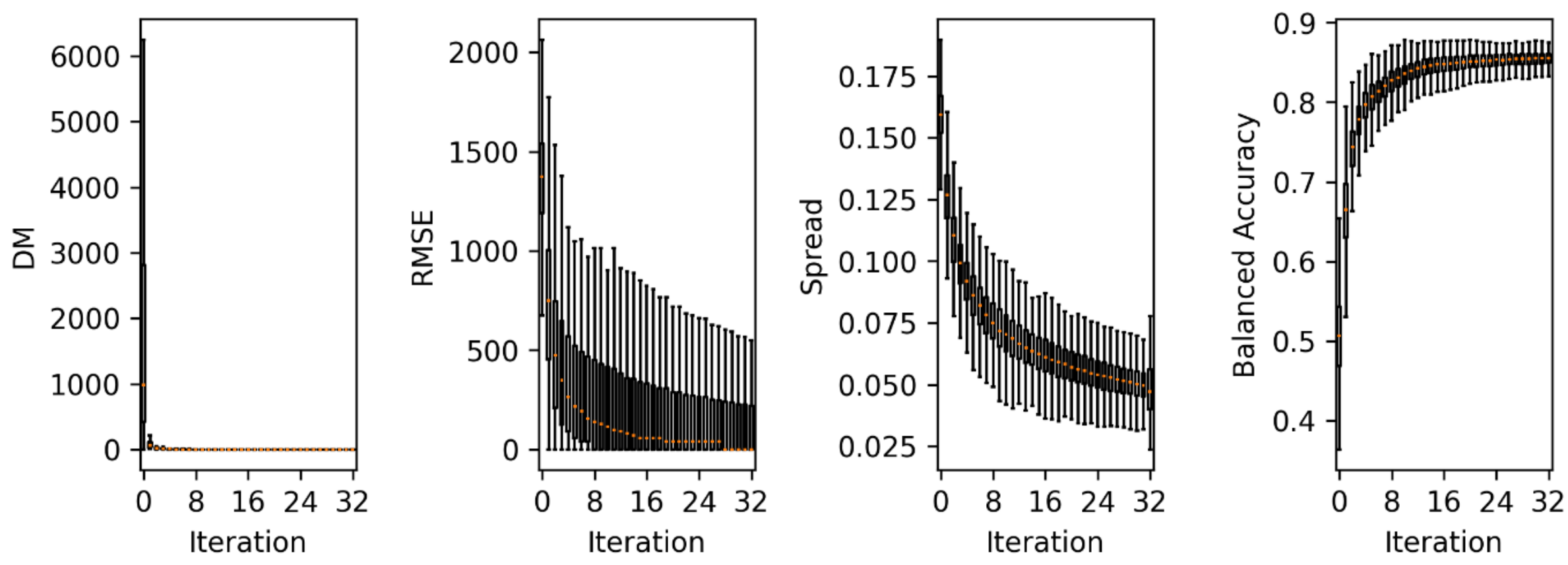

(b)

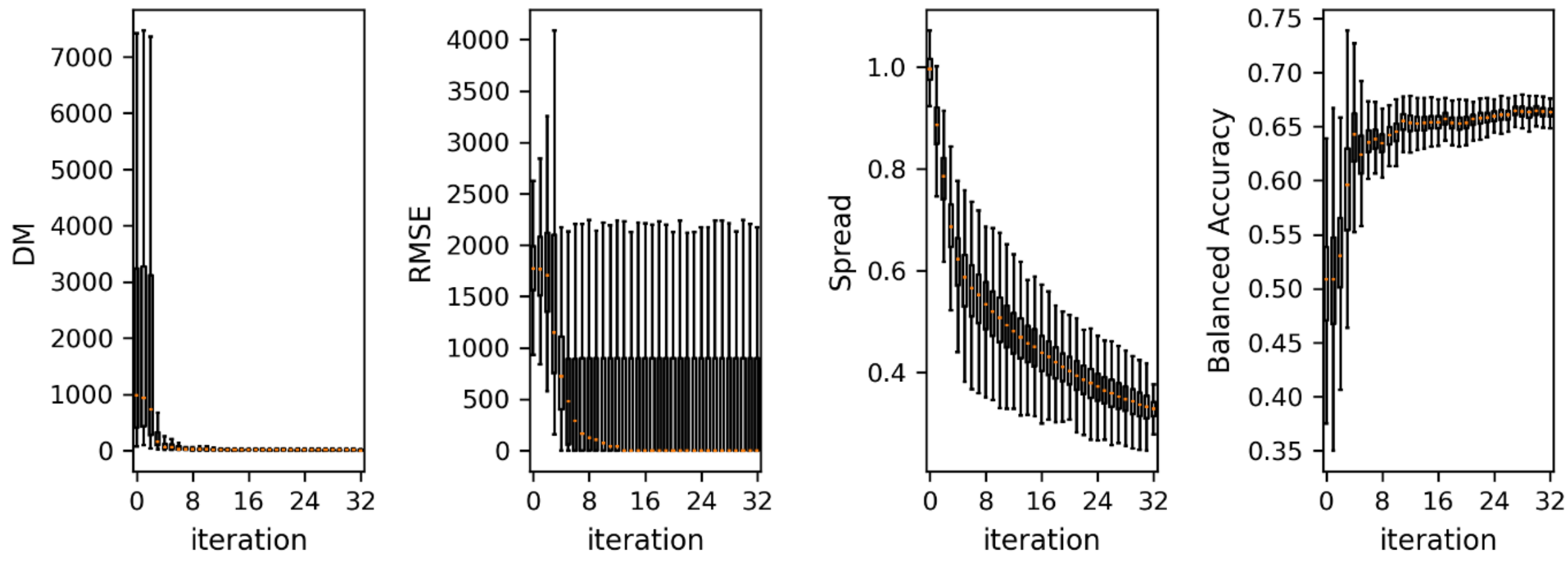

(c)

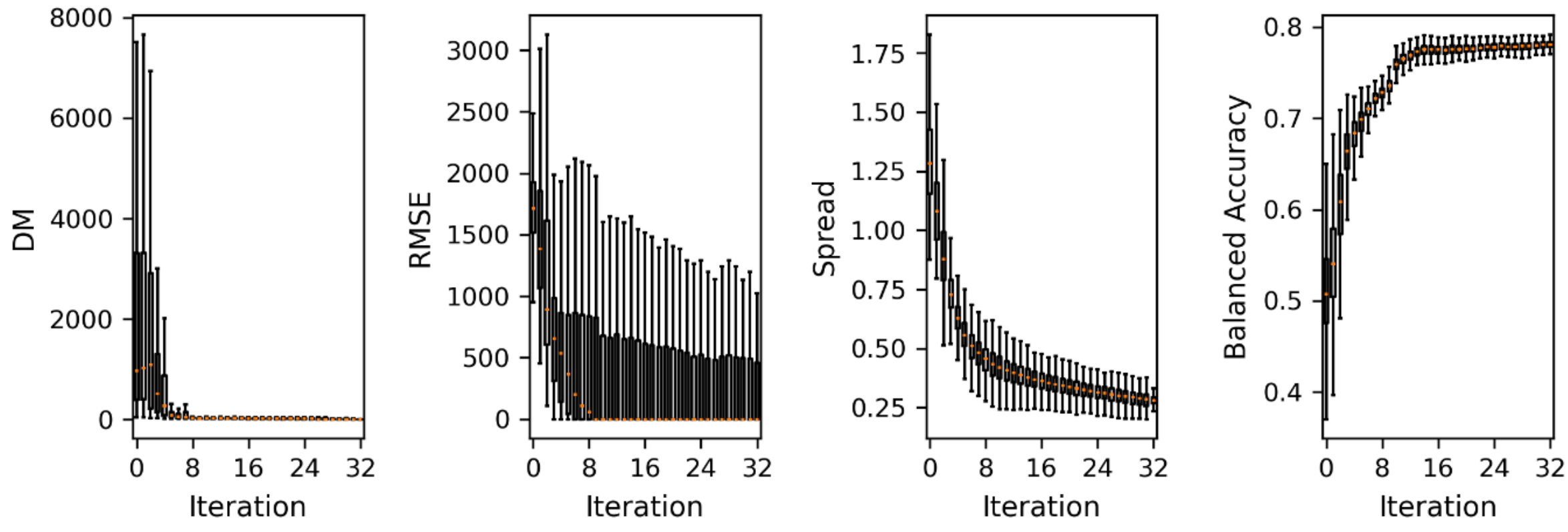

(d)

Figure 16: Graphs of data mismatching, RMSE, spread and balanced accuracy as a function of the number of iterations in the categorical case for: (a) VAE-GAN, (b) Latent Diffusion, (c) StyleGAN2 (z-space) and (d) StyleGAN2 (w-space).

Figure 17 shows boxplots for assimilation with StyleGAN2 in the z and w spaces, displayed together to facilitate comparison. As can be seen, assimilation in the w-space outperformed that in the z-space, confirming that the w-space is less entangled than the z-space, which contributes to better ESMDA performance.

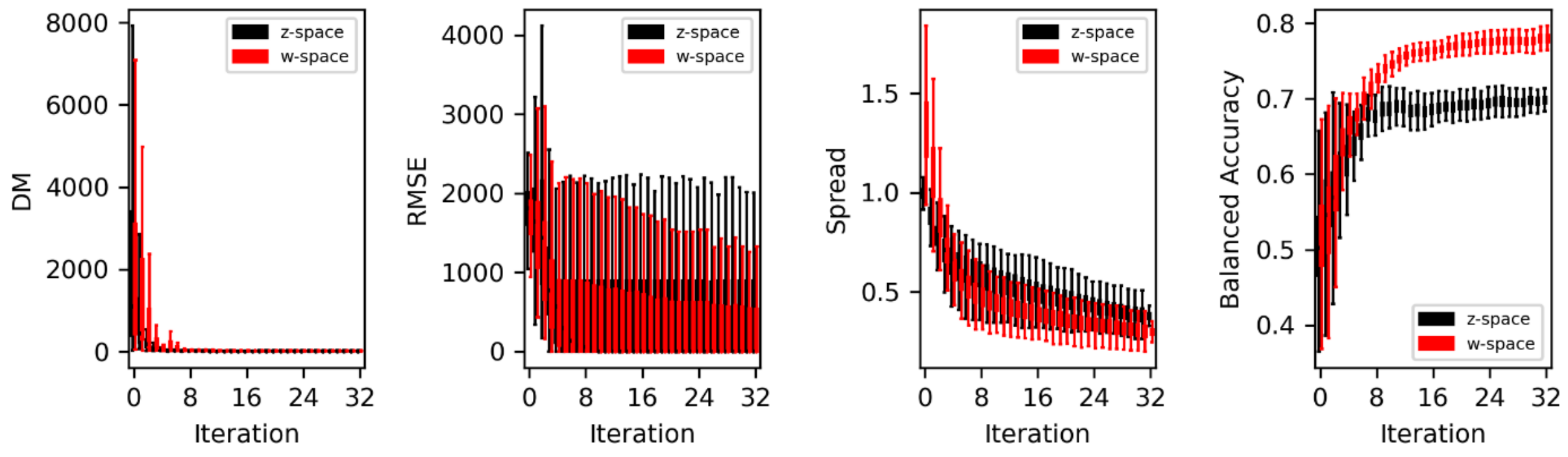

Figure 17: Comparison between graphs of data mismatching, RMSE, spread and balanced accuracy as a function of the number of iterations in the categorical case for StyleGAN2 (z-space) in black color and StyleGAN2 (w-space) in red color.

## 4.2. Case Study 2: Continuous training dataset

### *4.2.1. Training of Models*

***Variational Autoencoder Generative Adversarial Network (VAE-GAN)***

The VAE-GAN model was trained until the early stopping criterion was reached, achieving an FID of 876 and an FRD of 9.1 upon convergence, using 10,000 samples. Total training time was only 31 minutes. Figure 18 depicts the visual quality of the realizations reconstructed by the VAE-GAN model (generator), showing high-fidelity relative to the truth samples.

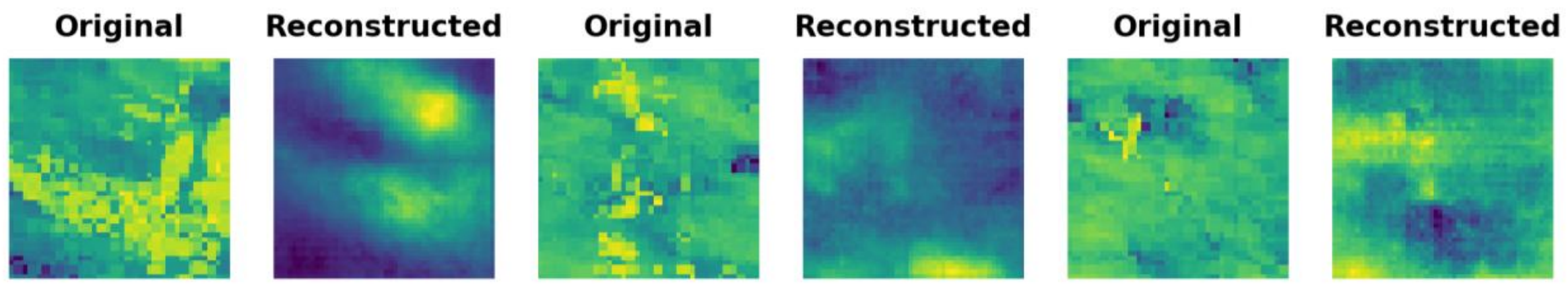

Figure 18: Three examples of random images from the set along with the reconstructed images by VAE-GAN in the continuous case.

Figure 19 shows the total error curves for the training and the validation sets, considering the reconstruction KL, generator, discriminator, and perceptual loss components. The curves show stabilization in errors after 80 epochs.

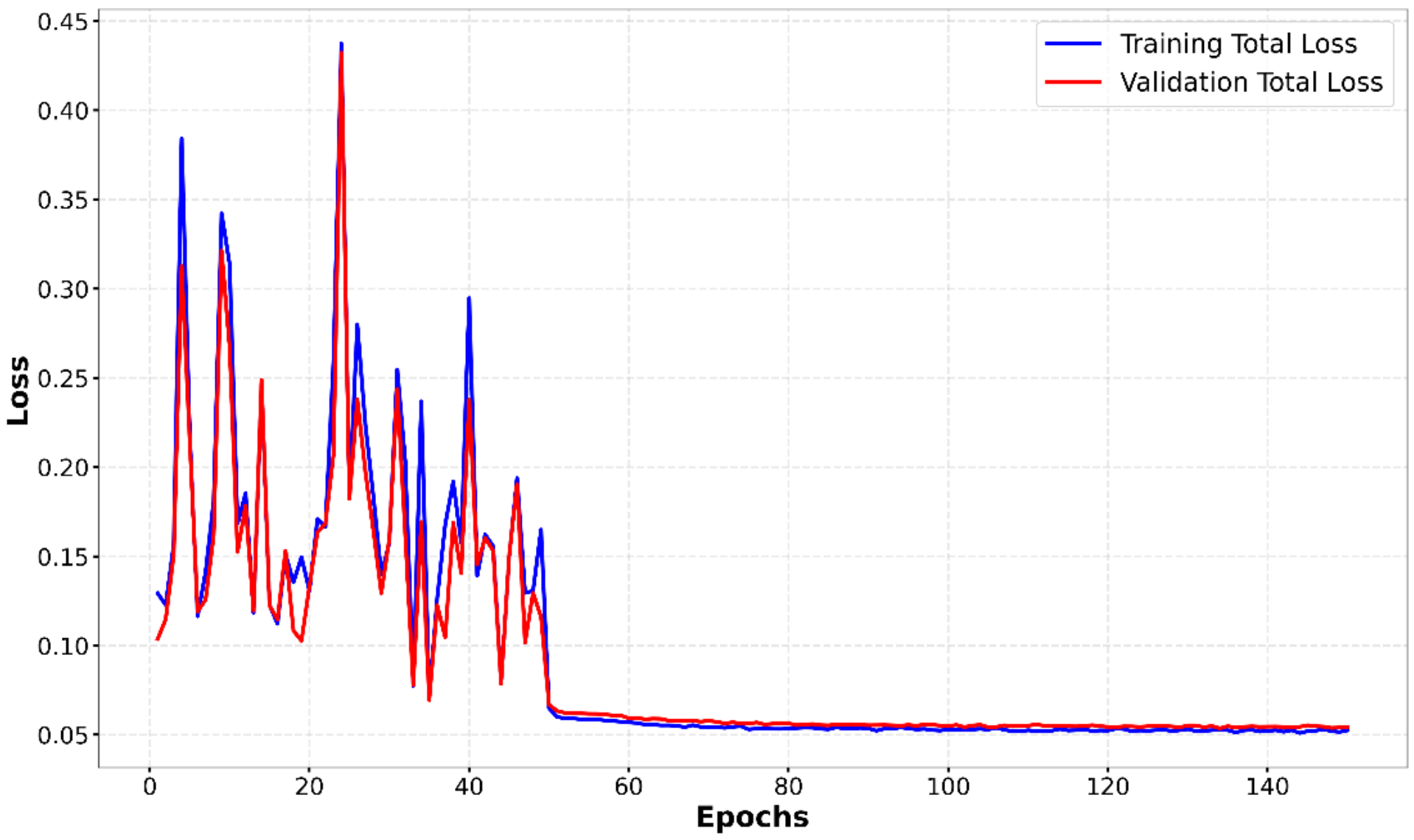

Figure 19: Total loss versus epochs with VAE-GAN in the continuous case.

### *Latent Diffusion*

We trained the LDM with continuous case after identifying the optimal architectural configuration until 133 epochs, reaching the early stopping criterion. The training time was 3 hours and 24 minutes, significantly increasing the computational time compared to the VAE-GAN model. Upon convergence, we obtained an FID and FRD, using 10,000 samples, equal to 167.6 and 5.4, respectively. This can be seen by the high-quality of the images generated at the end of the training, as shown in Figure 20.

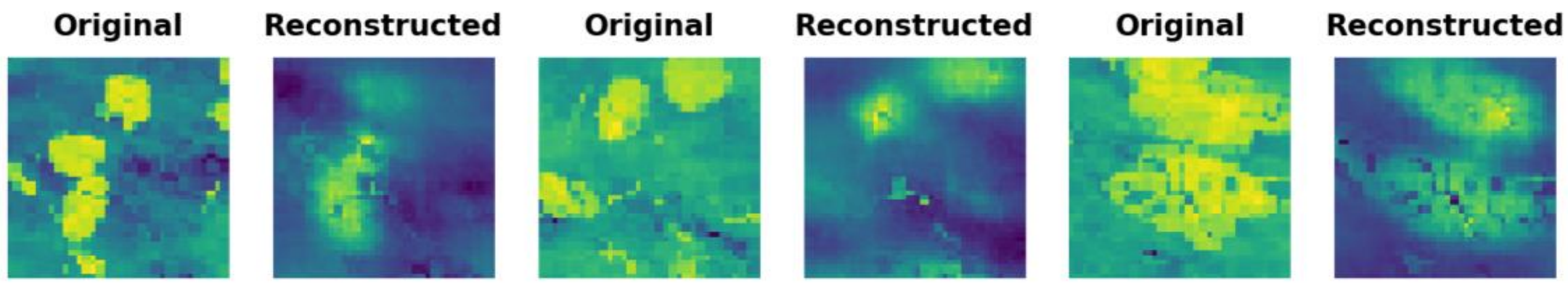

Figure 20: Images generated by LDM after training after 133 epochs in the continuous case.

Figure 21 shows the total diffusion loss throughout the training. As we can see, training and validation curves have decreased over time. Thus, we can see that in this case the training also was successful.

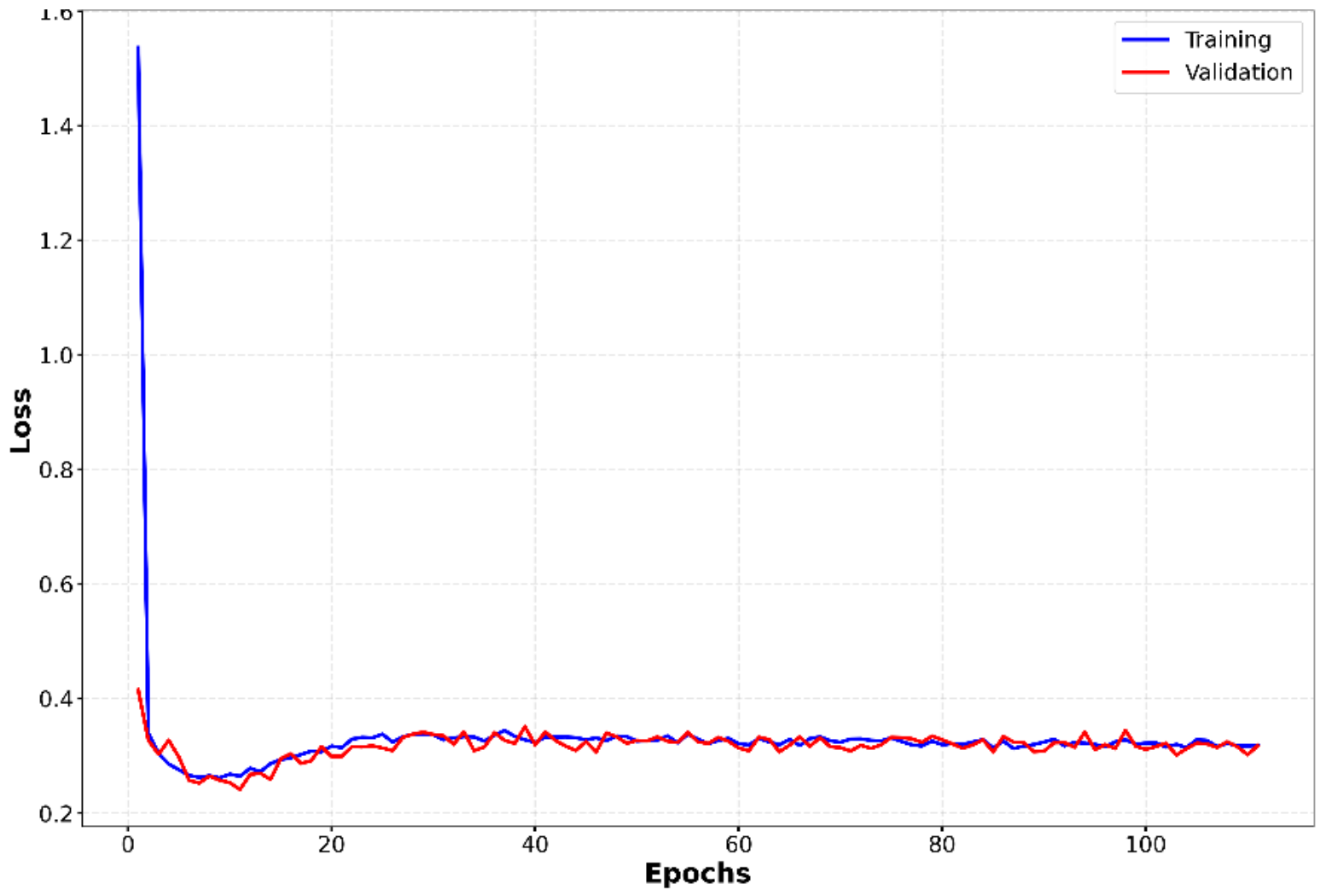

Figure 21: MSE of training and validation versus epochs with LDM across training iterations for case study 2.

***Style-Based Generative Adversarial Network (StyleGAN)***

The StyleGAN2 model was trained until reach 114 epochs, when the early stopping criterion was reached. We obtained the FID and FRD after the end of iterations, estimated using 10,000 samples, respectively equal to 513.4 and 4.6. The training time was only 58 minutes, and therefore, much faster than the LDM, but with almost twice the time of the VAE-GAN. We can see in Figure 22 the quality of the images generated by the generator's network, showing images with high-quality, similar to other models.

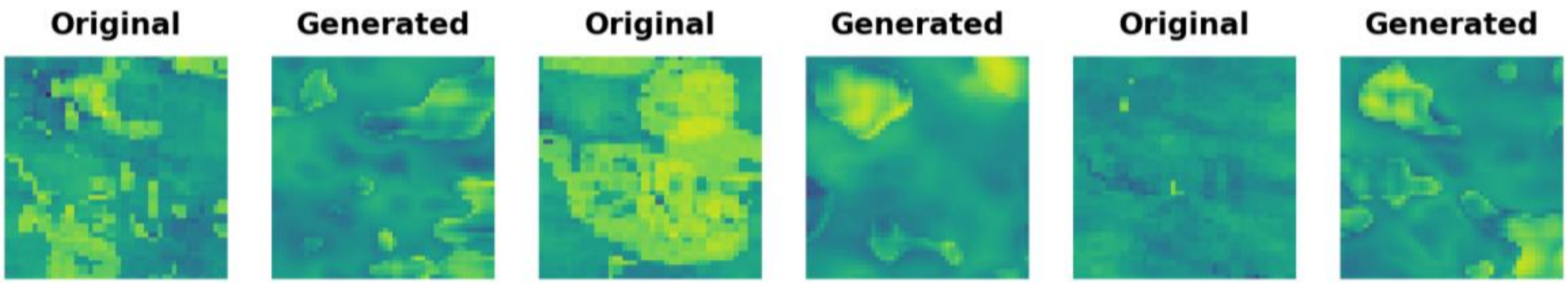

Figure 22: Three examples of random images using StyleGAN2 in the case study 2.

Figure 23 shows the MSE of the generator and discriminator. As we can see the fast decrease of generator loss and fast increase of discriminator throughout the training, showing that 114 epochs were sufficient to train the model.

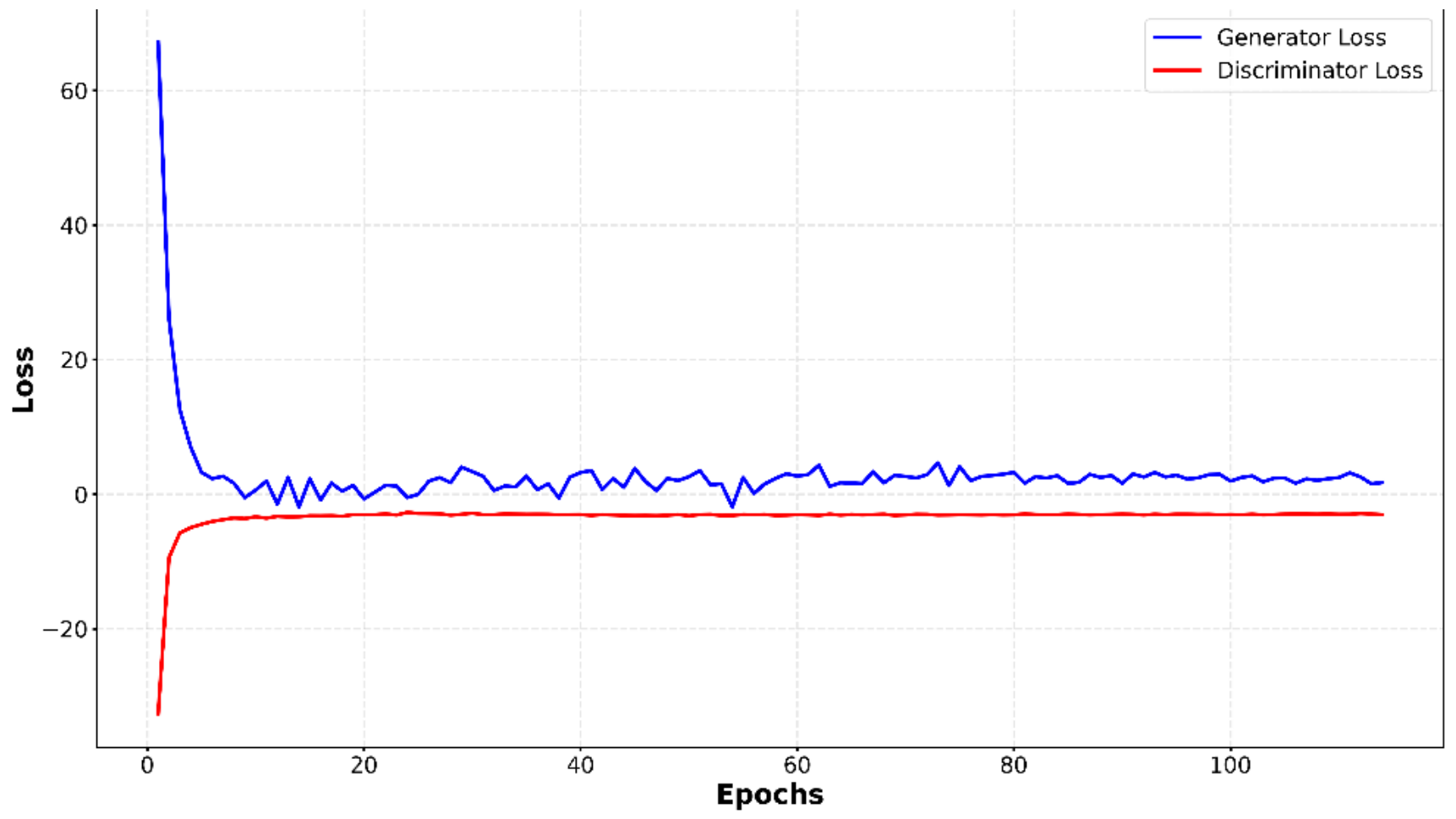

Figure 23: MSE of generator and discriminator losses versus epochs with StyleGAN2 in the case study 2.

In Table 3, we summarize the results of the FID and FRD metrics at the end of training for the three models. We can conclude again that the FRD metric was more appropriate than the FID also for this case study and that the values obtained were low, demonstrating that the training of all models was efficient, with the StyleGAN2 model achieving the lowest value.

Table 3: Results of FID and FRD of models in the continuous case.

| | FID | FRD |
|---|---|---|
| **VAE-GAN** | 876 | 9.1 |
| **Latent Diffusion** | 167.6 | 5.4 |
| **StyleGAN** | 513.4 | 4.6 |

In The three models show a different profile in the continuous case than in the discrete case, with StyleGAN now leading most metrics. StyleGAN is best on variogram MSE (0.00009), connectivity MSE (0.00374), histogram KL (0.15458), and MDS MMD (0.15714), suggesting it reproduces both spatial continuity and value distribution better than the alternatives in the continuous setting. Latent Diffusion remains intermediate, with a variogram MSE of 9.29604, connectivity MSE of 0.01020, histogram KL of 0.40680, and MDS MMD of 0.22051; its variogram error is still orders of magnitude larger than the other two models. VAE-GAN is worst on variogram MSE (0.00168 is actually its best metric relative to the others), connectivity MSE (0.05479), histogram KL (3.29511), and MDS MMD (0.48915), and its PCA correlation of 0.00027 is essentially zero. PCA correlation is very low for all three models (0.00027, 0.03229, 0.02959), making it uninformative in this continuous comparison as well, though Latent Diffusion and StyleGAN are marginally better than VAE-GAN. Overall, StyleGAN dominates the continuous case across four of five metrics, Latent Diffusion is second, and VAE-GAN is last.

Table 4, based on the comparative analysis of static geostatistical metrics. The three models show a different profile in the continuous case than in the discrete case, with StyleGAN now leading most metrics. StyleGAN is best on variogram MSE (0.00009), connectivity MSE (0.00374), histogram KL (0.15458), and MDS MMD (0.15714), suggesting it reproduces both spatial continuity and value distribution better than the alternatives in the continuous setting. 

Table 4: Results for static geostatistical metrics for the continuous case.

| | VAE-GAN | Latent Diffusion | StyleGAN |
|---|---|---|---|
| **Variogram MSE** | 0.00168 | 9.29604 | 0.00009 |
| **Connectivity MSE** | 0.05479 | 0.01020 | 0.00374 |
| **Histogram KL** | 3.29511 | 0.40680 | 0.15458 |
| **PCA Correlation** | 0.00027 | 0.03229 | 0.02959 |
| **MDS MMD** | 0.48915 | 0.22051 | 0.15714 |

### *4.2.2. Data Assimilation*

Figure 24 show the images of the true case, the priori and posteriori mean and standard deviation. We can observe that the posteriori mean presents an image similar to the true case for all models, but with more similarity in the cases of StyleGAN2 (z and w-spaces).

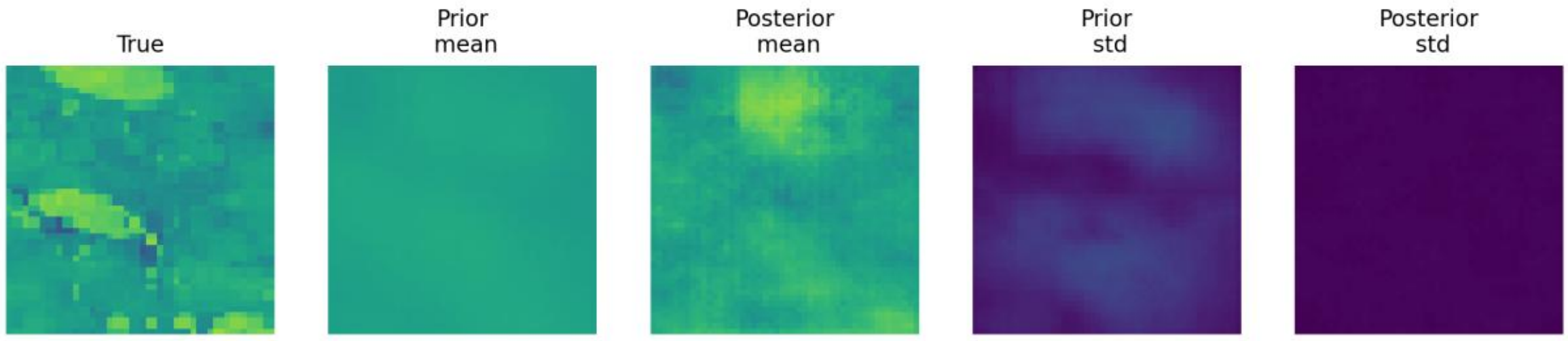

(a)

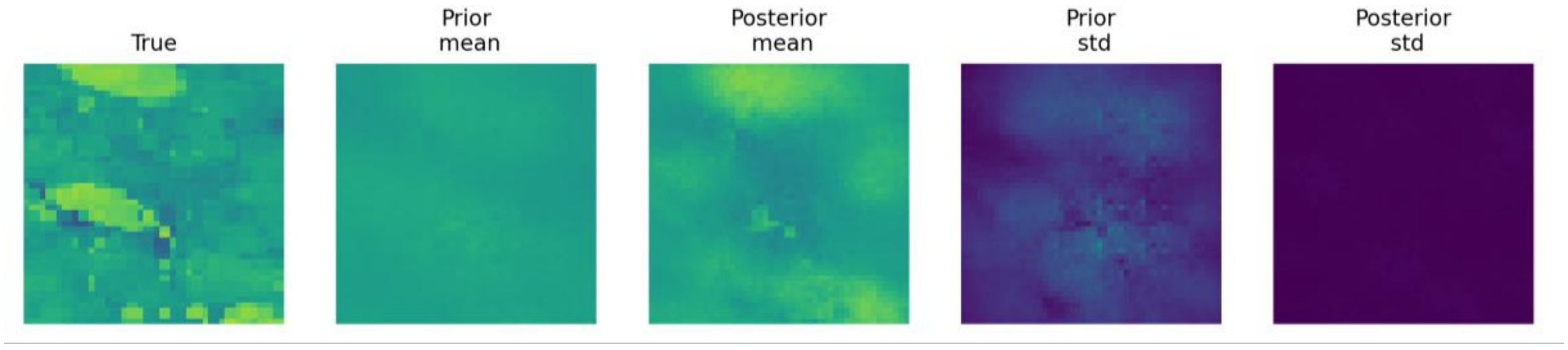

(b)

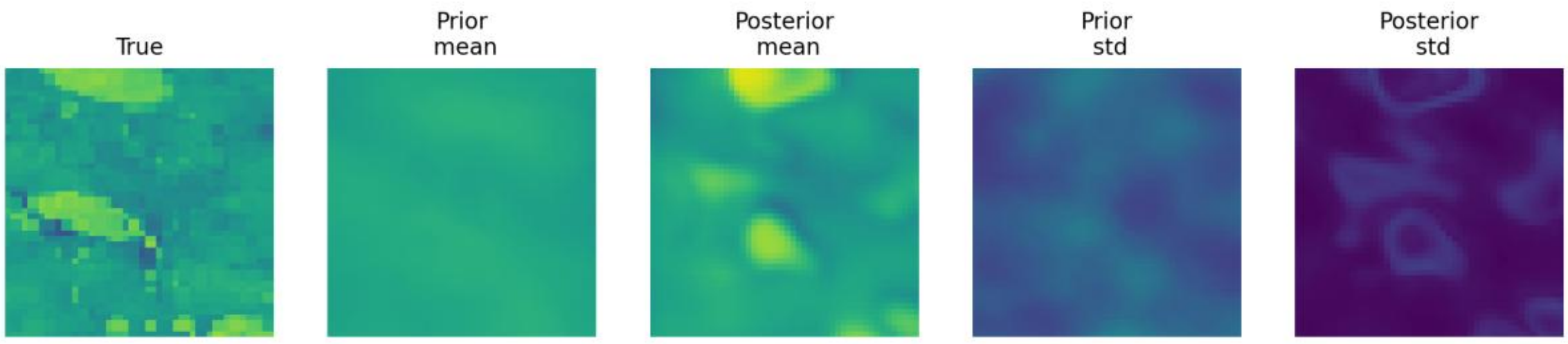

(c)

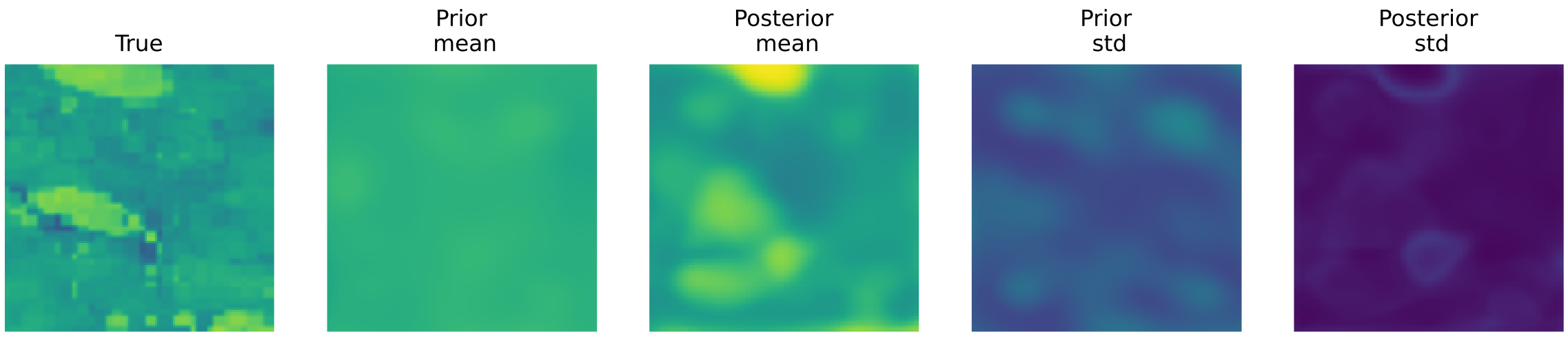

(d)

Figure 24: Images of the true case, the priori and posteriori mean, and the priori and posteriori standard deviation in the case study 2 for: (a) VAE-GAN, (b) LDM, (c) StyleGAN2 (z-space), and (d) StyleGAN2 (w-space).

Evaluating the time series of production data in the Figure 25, we can see a better matching for the LDM and StyleGAN2 in the w-space than VAE-GAN and StyleGAN2 in the z-space. We can clearly see that the assimilation of StyleGAN2 using the intermediate latent space w was better than with the latent space z.

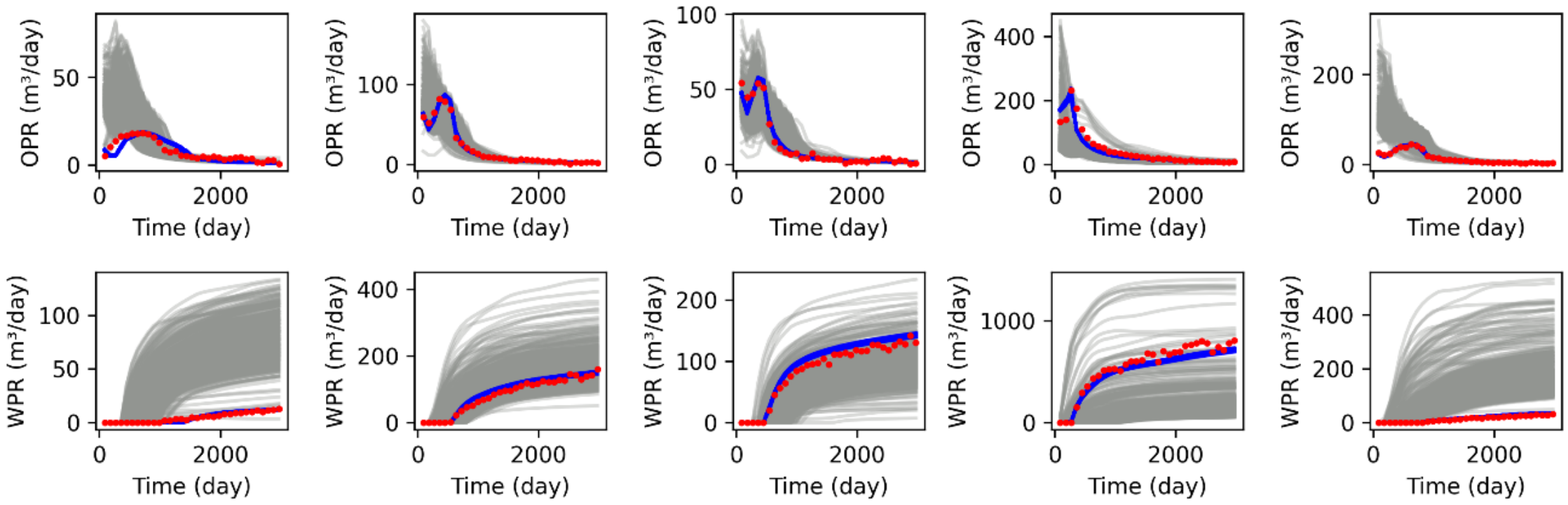

(a)

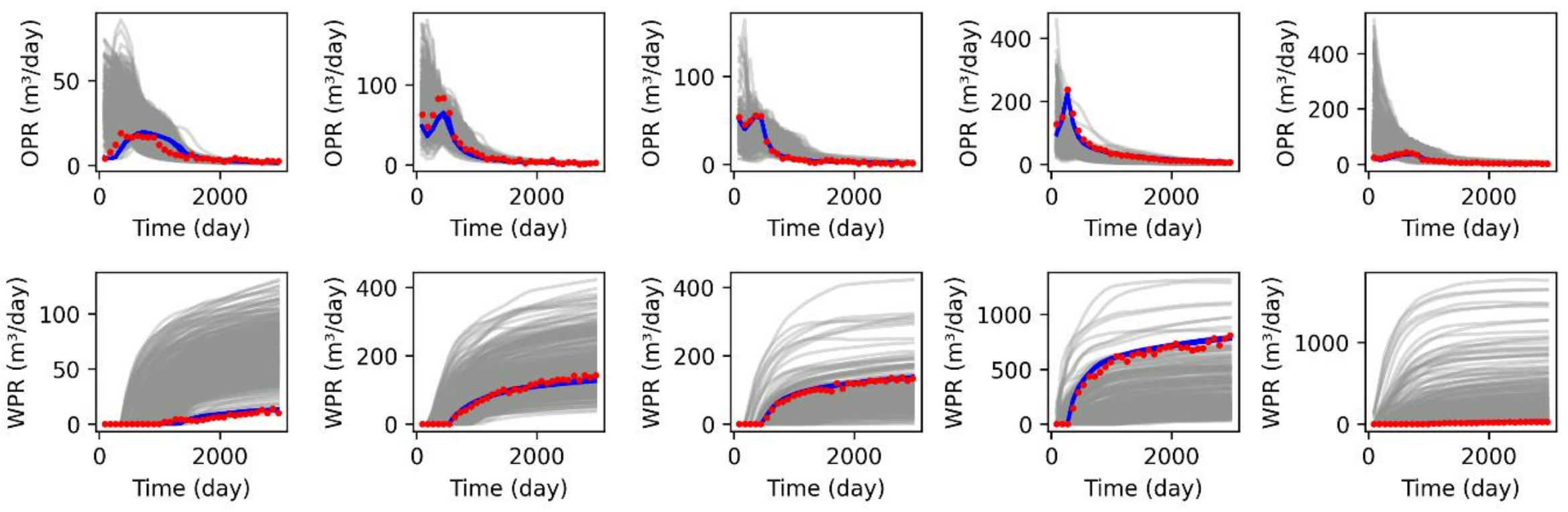

(b)

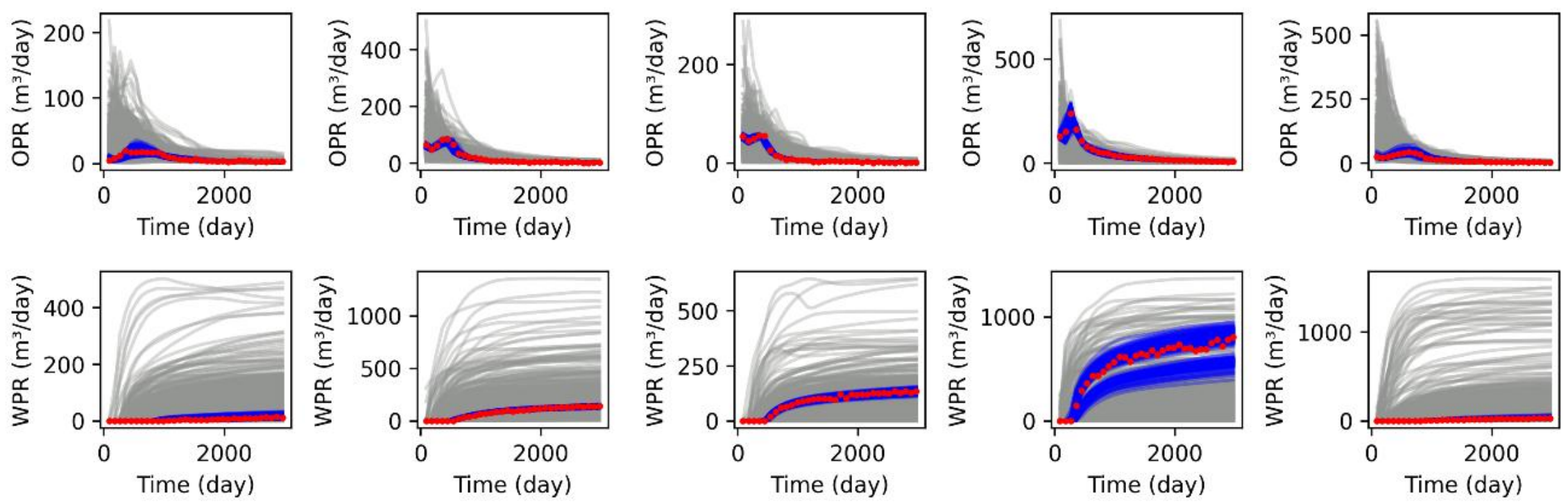

(c)

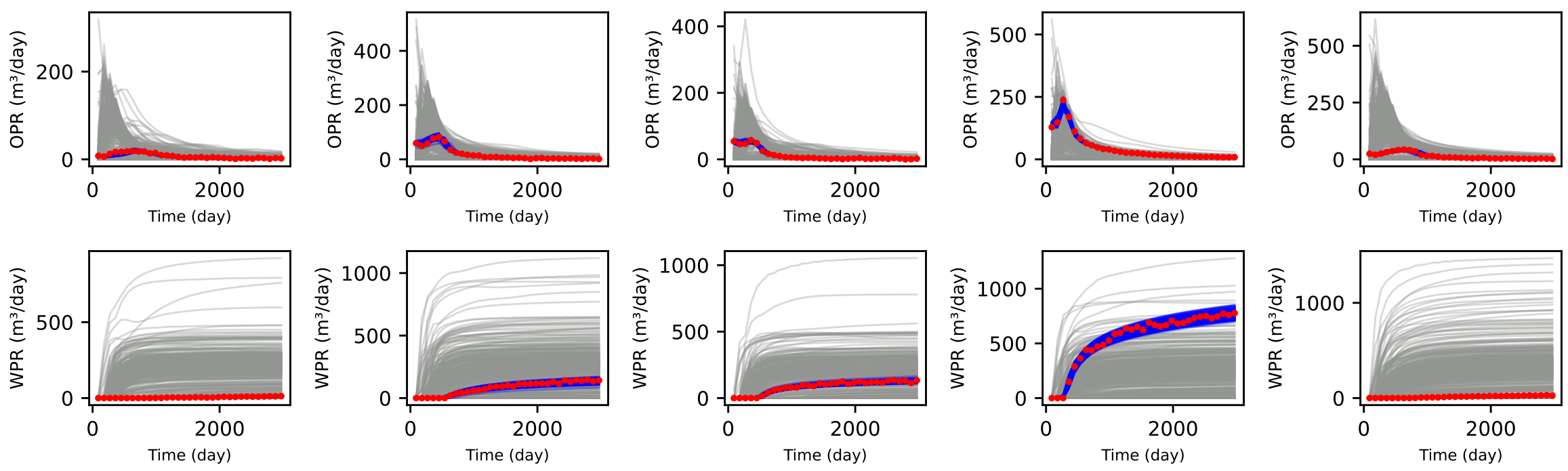

(d)

Figure 25: Time series of production data from the first five producers of oil production rate (above) and water production rate (below). Here, the gray lines represent the prior ensemble, blue lines represent the posterior ensemble, and the red dots represent the measurements in the case study 2 for: (a) VAE-GAN, (b) LDM, (c) StyleGAN2 (z-space) and (d) StyleGAN2 (w-space).

Figure 26 shows graphs of data mismatching, RMSE, spread as a function of the number of iterations. We can observe that LDM and StyleGAN2 in the w-space obtained the best results for RMSE and spread, followed by StyleGAN2 in the z-space and VAE-GAN, in this sequence.

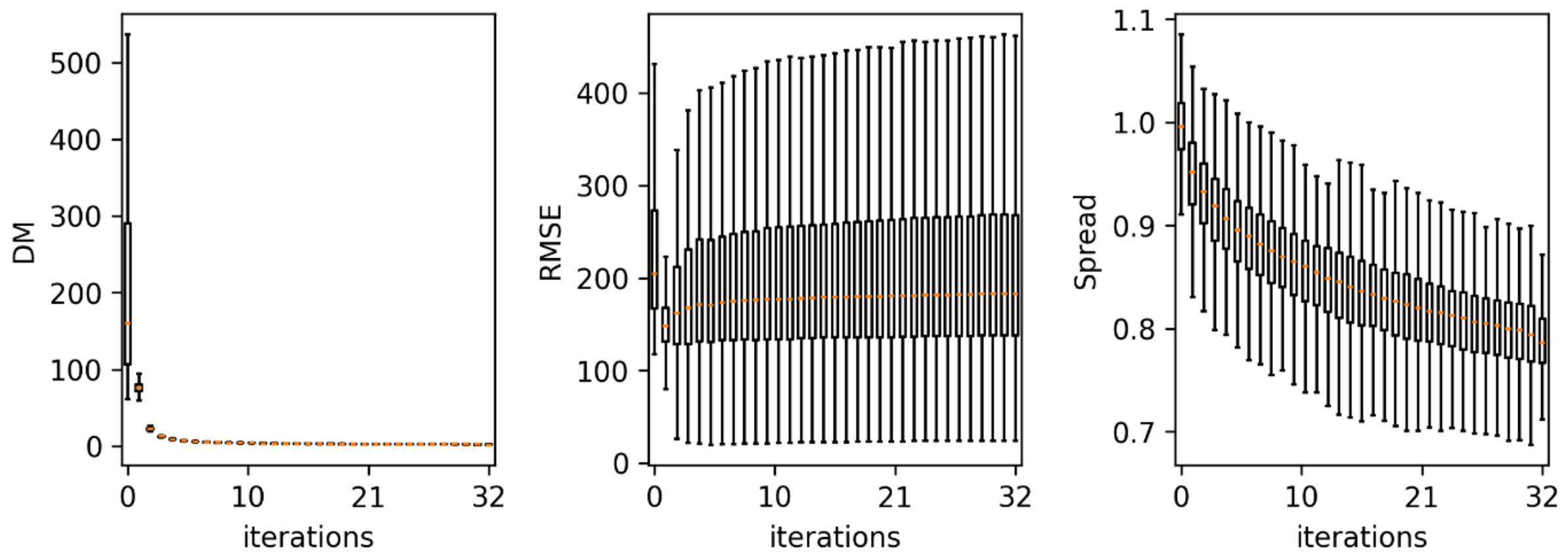

(a)

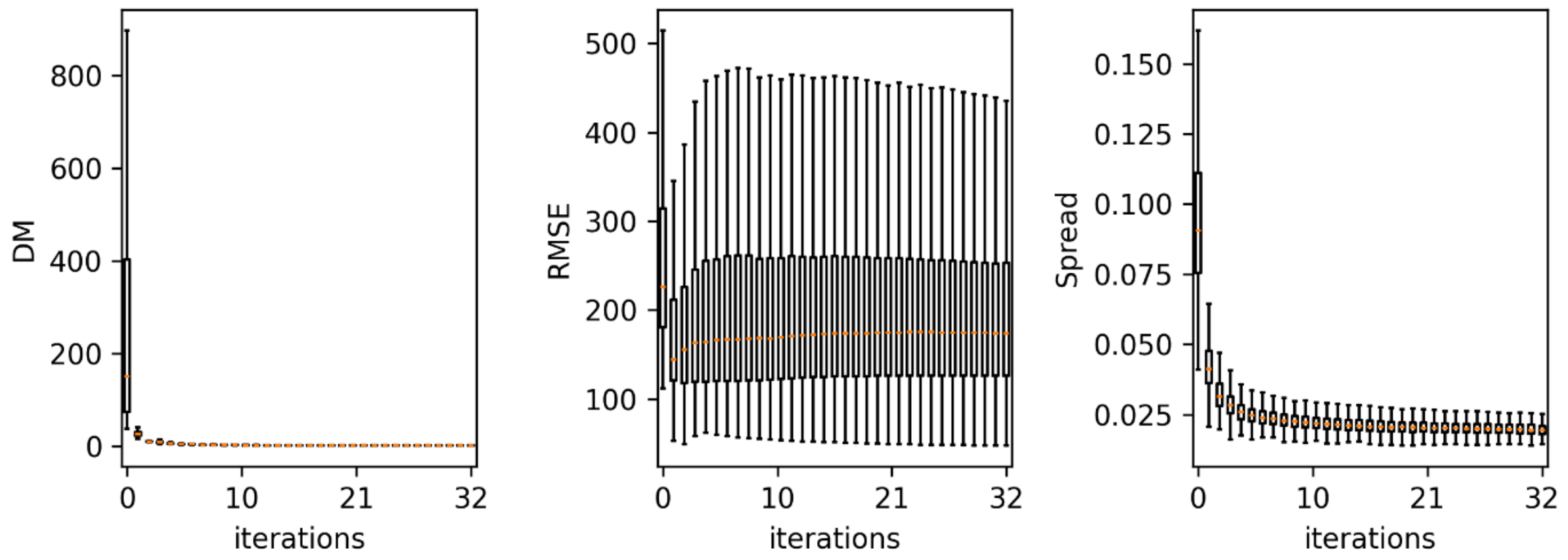

(b)

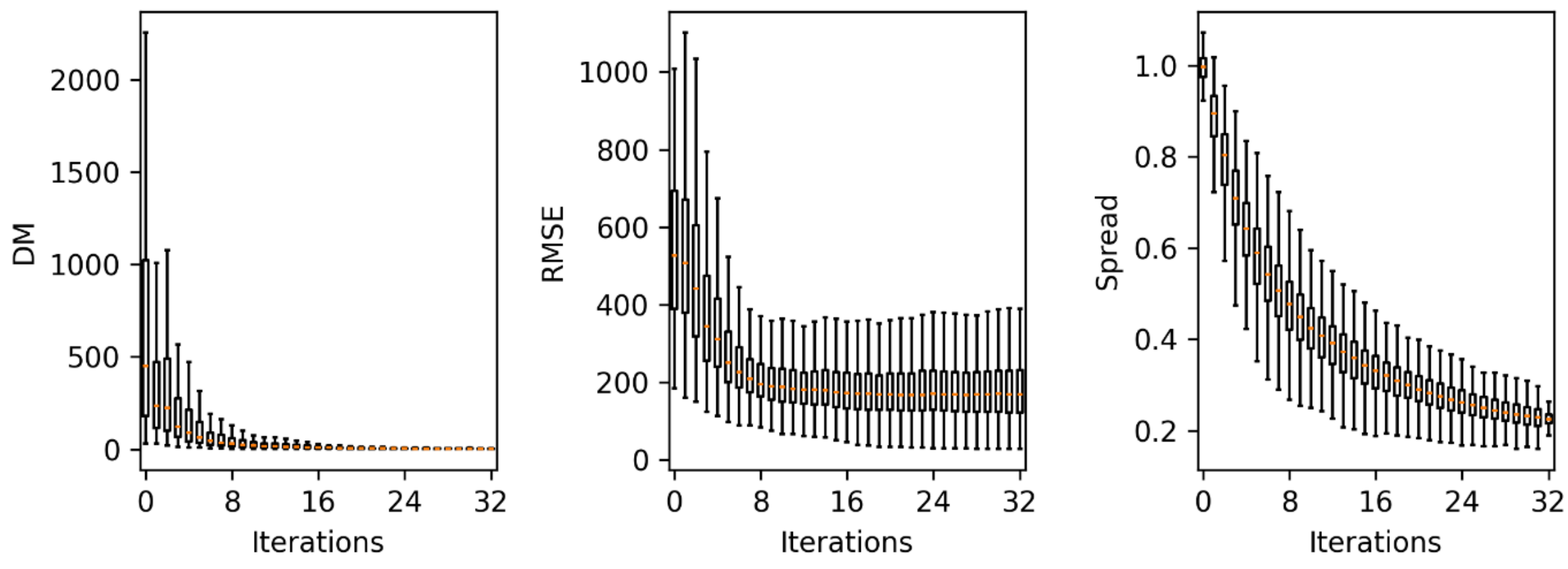

(c)

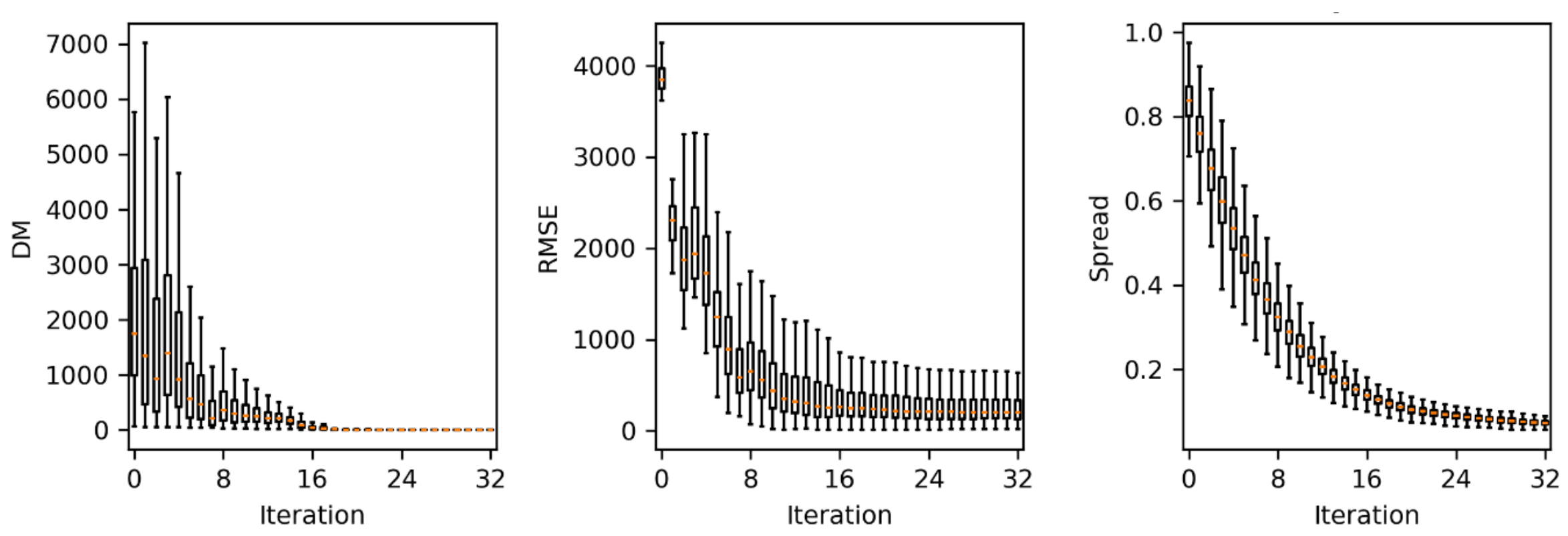

(d)

Figure 26: Graphs of data mismatching, RMSE and spread as a function of the number of iterations for: (a) VAE-GAN, (b) LDM, (c) StyleGAN2 (z-space) and (d) StyleGAN2 (w-space).

Figure 27 shows boxplots for assimilation with StyleGAN2 in the z and w spaces, displayed together. As we can see again, assimilation in the w-space outperformed that in the z-space, confirming that the w-space is less entangled than the z-space, which contributes to better DA efficiency.

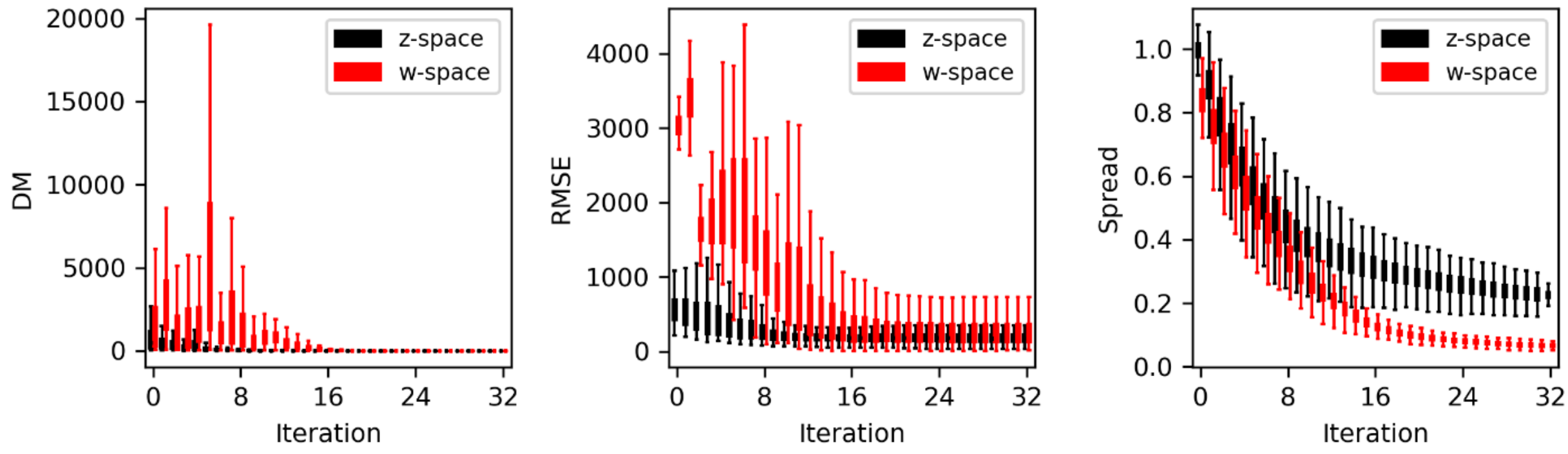

Figure 27: Comparison between graphs of data mismatching, RMSE, spread as a function of the number of iterations in the continuous case for StyleGAN2 (z-space) in black color and StyleGAN2 (w-space) in red color.

## 5. CONCLUSIONS

This work shows that StyleGAN2 generates high-quality realizations with high training stability, driven by its disentangled latent space that is fundamental to a high performance of data assimilation using ESMDA. The results based on geostatistical metrics demonstrated the superior quality of the StyleGAN2 model in generating high-quality samples with geological realism. Our findings show that performing data assimilation with StyleGAN2 using the intermediate space ($w$-space) yielded better results than the traditional formulation which uses the latent space (z-space). This is due to the fact that ESMDA uses linear updates and the w-space is much more linear and disentangled than the highly entangled z-space. And since the w-space has already been mapped by the network, the updated vectors remain close to realistic geological patterns. Our results were compared against those of two models considered state-of-the-art in recent literature, VAE-GAN and LDM. The results showed that StyleGAN offers a competitive computational cost compared to other models, yielding an excellent matching, producing samples with more geological realism than other models. The results of DA were similar between LDM and StyleGAN2, particularly when the latter uses the intermediate latent space.

Future research could focus on evaluating the proposed approach on three-dimensional, large-scale reservoir models, as well as investigating techniques to mitigate spurious correlations (such as inflation and localization methods) when using these deep generative frameworks for parameterization in ensemble-based data assimilation.

## Acknowledgments

We gratefully acknowledge the support of Escola Politécnica of the University of São Paulo and Imperial. The authors would also like to thank the LASG (Laboratory of Reservoir Simulation and Management) for supporting this research, CMG (Computer Modelling Group Ltd.) for providing the reservoir simulator licenses used in this study and FAPESP – São Paulo Research Foundation (16/08801-0). We also thank the National Council for Scientific and Technological Development (CNPq) for financial support through grant number 310676/2025-8.

## Conflicts of interest

The authors declare that they have no known competing financial interests or personal relationships that could have appeared to influence the work reported in this paper.

## Computer Code Availability

The codes and the datasets generated and/or analyzed during the current study are available in the Github repository: https://github.com/LASG-USP/Parameterization_StyleGAN.